\documentclass{ieeeaccess}
\usepackage{cite}
\usepackage{amsmath,amssymb,amsfonts}
\usepackage{algorithmic}
\usepackage{algorithm}
\usepackage{graphicx}
\usepackage{textcomp}
\usepackage{tabularx}
\usepackage{microtype}
\usepackage[caption=false,font=footnotesize]{subfig}
\usepackage{multirow}
\usepackage{booktabs}
\usepackage{bm}
\usepackage{url}
\usepackage[dvipsnames,table,xcdraw]{xcolor}
\usepackage{pifont} 
\usepackage{array}
\newcommand{\rev}[1]{\textcolor{black}{#1}}
\newenvironment{revblock}{\begingroup\color{black}}{\endgroup}

\DeclareGraphicsExtensions{.pdf,.png,.jpg,.jpeg}
\graphicspath{{Figure/}{author/}}

\begin{document}

\history{Date of publication xxxx, date of current version xxxx.}
\doi{10.1109/ACCESS.2025.XXXXXXX}
\title{PENet+: A Lightweight Residual Transformer Framework for Efficient Image Steganalysis}

\author{
\uppercase{JINCHEOL AN}\authorrefmark{1,3},
\uppercase{Dongsu KIM}\authorrefmark{2},
\uppercase{Haneol Jang}\authorrefmark{2},
and \uppercase{YOUNGJOON YOO$^{\dagger}$}\authorrefmark{1,3}}

\address[1]{Department of Artificial Intelligence, Chung-Ang University, Seoul, Korea (e-mail: \{wlscjf2496, yjyoo3312\}@cau.ac.kr)}
\address[2]{Hanbat National University, Department of Computer Engineering, 125, Dongseo-daero, Yuseong-gu, Daejeon, 34158, Korea (e-mail: 30251264@edu.hanbat.ac.kr, hejang@hanbat.ac.kr)}
\address[3]{SNUAILAB, Seoul, Korea}

\markboth
{An \textit{et al.}: PENet+: A Lightweight Residual Transformer Framework for Efficient Image Steganalysis}
{An \textit{et al.}: PENet+: A Lightweight Residual Transformer Framework for Efficient Image Steganalysis}

\corresp{Corresponding author: YoungJoon Yoo (e-mail: yjyoo3312@cau.ac.kr).}

\begin{abstract}

Image steganalysis, the detection of hidden information embedded in digital images, is a core component of modern cybersecurity and digital forensics. Recent residual Transformer architectures, such as Pixel-Difference-Convolution and Enhanced-Transformer-Network (PENet)~\cite{penet2023}, achieve strong detection accuracy, but their computational and memory demands hinder deployment in resource-constrained settings. In this paper, we present \textbf{PENet+}, a lightweight steganalysis framework that preserves PENet’s discriminative structure while substantially improving efficiency. Instead of redesigning or compressing the attention blocks, we retain PENet’s self-attention topology for reproducibility 
and introduce a classifier-streamlining stage that progressively narrows the SPP$\rightarrow$FC1 input channels, yielding large reductions in parameters and FLOPs with negligible accuracy loss.
\rev{Here, SPP denotes spatial pyramid pooling, FC1 denotes the first fully connected layer after SPP, and HPF denotes high-pass filter.}
We further refine the HPF stem with an activation-aware selection mechanism that aggregates high-pass filter responses in an early stage and selects a balanced SRM–Gabor top-$K$ subset, and we replace PENet’s backbone with a MobileNetV2-style inverted residual network for efficient spatial processing. A balanced HPF configuration with \mbox{$K{=}31$} filters (16 Gabor + 15 SRM) matches or surpasses the accuracy of heavier configurations while lowering compute. Finally, we revisit activation design and motivate PReLU from a steganalysis standpoint, arguing that preserving negative responses is beneficial for capturing weak stego cues that ReLU would otherwise suppress. 

On our disjoint ALASKA2 JPEG QF90 evaluation protocol at $512{\times}512$ resolution, which uses 5{,}000 cover images for training, validation, and internal testing and a separate 19{,}000-cover evaluation set, PENet+ achieves up to a 45.5\% reduction in parameters and approximately 97\% fewer FLOPs compared with the re-evaluated PENet baseline. These results suggest that the proposed design provides a computationally efficient direction for resource-constrained image steganalysis, although actual device-level latency and power measurements remain future work.
\end{abstract}


\titlepgskip=-15pt
\maketitle

\section{Introduction}\label{sec:intro}

Image steganalysis, the detection of hidden information in seemingly benign images, is a core primitive in modern security and digital forensics. It enables law enforcement to monitor covert channels, helps intelligence agencies uncover coordinated disinformation, and allows data-loss-prevention (DLP) systems to block image-based exfiltration of sensitive data. As encrypted messaging proliferates and AI-generated imagery becomes ubiquitous, robust steganalysis is simultaneously more challenging and more critical.

Traditional steganalysis relied on handcrafted statistical descriptors (e.g., co-occurrence matrices and noise-residual histograms) that are interpretable but struggle to generalize across diverse image distributions and modern embedding schemes. Deep models, especially convolutional neural networks (CNNs) and Transformers, have greatly improved robustness by capturing both local and global dependencies~\cite{qian2015deep,ye2017yenet,xu2016xunet,yedroudj2018cnn,srnet2018boroumand,tsang2018sid,you2021siastegnet,dosovitskiy2021vit,liu2021swin}. Pixel Difference Convolution and Enhanced Transformer Network (PENet)~\cite{penet2023} is a representative example, combining residual cue extraction with self-attention and achieving state-of-the-art accuracy on benchmarks such as ALASKA2.

However, PENet’s high accuracy comes with a significant computational cost. Although its parameter count ($\approx 16.16\text{M}$) is moderate, 
\rev{high-resolution convolutional processing, the large activation maps produced by the HPF stem, and the attention module together increase computational and memory burdens at $512\times512$ resolution.}
Consequently, deployment on resource-constrained platforms requires reducing the forward computational burden while preserving detection accuracy.


To address these constraints, we propose \textbf{PENet+}, a lightweight steganalysis architecture. 
First, we refine PENet into a lightweight variant designed to reduce computational cost at $512\times512$ resolution.
Second, we show that PENet+ preserves detection performance—measured by per-image accuracy and \rev{area under the ROC curve (AUC)} on the ALASKA2 QF90 benchmark—without sacrificing classification reliability. 
Achieving both properties makes the model more suitable for resource-constrained deployment than the original PENet, while actual device-level validation is left for future work.

 To summarize, our proposed \textit{PENet+} bridges high-accuracy steganalysis and efficient edge deployment through classifier-centric compression, activation-aware design, and activation-aware HPF selection:

\begin{itemize}
\item \textbf{Activation-aware HPF selection.} Instead of using all 62 fixed filters, we estimate per-filter importance from HPF activations (either via warm-up statistics or a learnable selector network) and select a balanced SRM--Gabor top-$K$ subset, yielding a compact and informative HPF stem with stable convergence.
\item \textbf{Classifier bottleneck streamlining (small, complementary).} We reduce the SPP$\rightarrow$FC1 input dimension $D_{\text{in}} = C\cdot S$ by narrowing the channel width $C$ before SPP (e.g., $C\!\in\!\{32,64,128\}$), yielding additional savings with negligible accuracy loss.
\item \textbf{Domain-motivated PReLU.} Preserving negative activations helps capture weak stego cues; we therefore adopt PReLU with per-channel learnable slopes, with the gains most pronounced in compact classifiers.
\item \textbf{Lightweight backbone with inverted residuals.} MobileNetV2-style blocks replace heavy stacks to cut spatial FLOPs while preserving residual sensitivity.
\item \textbf{Reproducible attention.} For fair comparison, we keep PENet’s self-attention topology unchanged; efficiency gains mainly arise from the backbone and HPF selection, with classifier streamlining as a complement.
\end{itemize}

These contributions demonstrate that \textit{PENet+} achieves an effective trade-off between structural simplification and discriminative optimization while maintaining robustness across multiple steganographic algorithms, providing a practical and deployable steganalysis framework for resource-limited environments.

\section{Related Works}

\subsection{CNN-based Steganalysis}
Early attempts such as Qian et al.'s CNN-based steganalyzer~\cite{qian2015deep} demonstrated that deep networks could automatically learn discriminative residual patterns for steganalysis. YeNet~\cite{ye2017yenet} introduced the first CNN that surpassed rich models~\cite{fridrich2012rich} by learning spatial dependencies directly from noise residuals. 
XuNet~\cite{xu2016xunet} and Yedroudj-Net~\cite{yedroudj2018cnn} further incorporated batch normalization and \rev{truncated linear unit (TLU)} activation to improve convergence.
SRNet~\cite{srnet2018boroumand} deepened this paradigm with hierarchical residual connections, achieving strong generalization across multiple embedding algorithms. However, CNN-based architectures primarily rely on local receptive fields and therefore struggle to model long-range spatial relationships, a limitation that attention-based designs aim to alleviate.

\subsection{Transformer-based Steganalysis}
Recent works have explored Transformer architectures to better capture global dependencies. Vision Transformers (ViT)~\cite{dosovitskiy2021vit} and Swin Transformers~\cite{liu2021swin} demonstrated the effectiveness of self-attention for long-range reasoning, inspiring their integration into steganalysis pipelines. PENet~\cite{penet2023} combines pixel-difference convolution (PDConv) with multi-head self attention (MHSA).  
To ensure fair comparison with prior work, many Transformer-based systems preserve full-resolution feature maps until late stages. However, attention complexity grows quadratically with the number of tokens ($T = HW$), making MHSA particularly expensive for $512{\times}512$ images widely used in steganalysis.

\subsection{Lightweight Architecture Design}
Lightweight models such as MobileNetV2~\cite{sandler2018mobilenetv2}, ShuffleNet~\cite{zhang2018shufflenet}, and EfficientNet~\cite{tan2019efficientnet} achieve favorable accuracy--efficiency trade-offs through depthwise separable convolutions and inverted residual blocks. These strategies have inspired efficient forensic pipelines such as deepfake detection and manipulation localization~\cite{li2018gcpnorm,deng2019gcpsteganalysis,zhang2020depthwise}.  
However, steganalysis imposes additional challenges because stego artifacts exist at extremely low signal-to-noise ratios. PDConv-based or early-stage residual-enhancement blocks must maintain high sensitivity while reducing computation, a non-trivial trade-off that limits the applicability of generic lightweight designs.

\rev{In addition to lightweight backbone design, early residual redundancy remains an important issue for efficient steganalysis. Existing CNN based steganalyzers often use fixed residual filters or wide early features to preserve weak stego cues~\cite{ye2017yenet,yedroudj2018cnn,srnet2018boroumand,penet2023}, but these designs can increase memory and computation by propagating redundant responses at high resolution. This motivates our HPF selection strategy, which keeps a compact but informative residual front end.}

\rev{Recent steganalysis studies improve efficiency using lightweight CNN backbones, EfficientNet based pipelines, and large scale pretraining for JPEG steganalysis~\cite{butora2021pretrain,yousfi2020imagenetjpeg,yousfi2021improvingeffnet}. 
Unlike these CNN based methods, PENet+ improves residual Transformer efficiency through HPF redundancy reduction, an inverted residual backbone, and SPP to FC streamlining while preserving the attention topology.}

\subsection{Activation Function Studies}
Activation functions greatly influence gradient flow and residual sensitivity.
Early steganalysis networks relied on ReLU-like activations for numerical stability. Later variants such as ReLU6~\cite{howard2017mobilenets}, LeakyReLU~\cite{maas2013rectifier}, and SiLU~\cite{elfwing2018silu} were explored to reduce saturation and preserve negative information. PReLU~\cite{he2015prelu} introduced learnable negative slopes, improving convergence in low-level vision tasks. 
\rev{Prior steganalysis work also reports benefits of retaining small negative responses for subtle-noise detection~\cite{ye2017yenet,srnet2018boroumand}.}

\begin{figure*}[!t]
  \centering
  \includegraphics[width=0.95\linewidth]{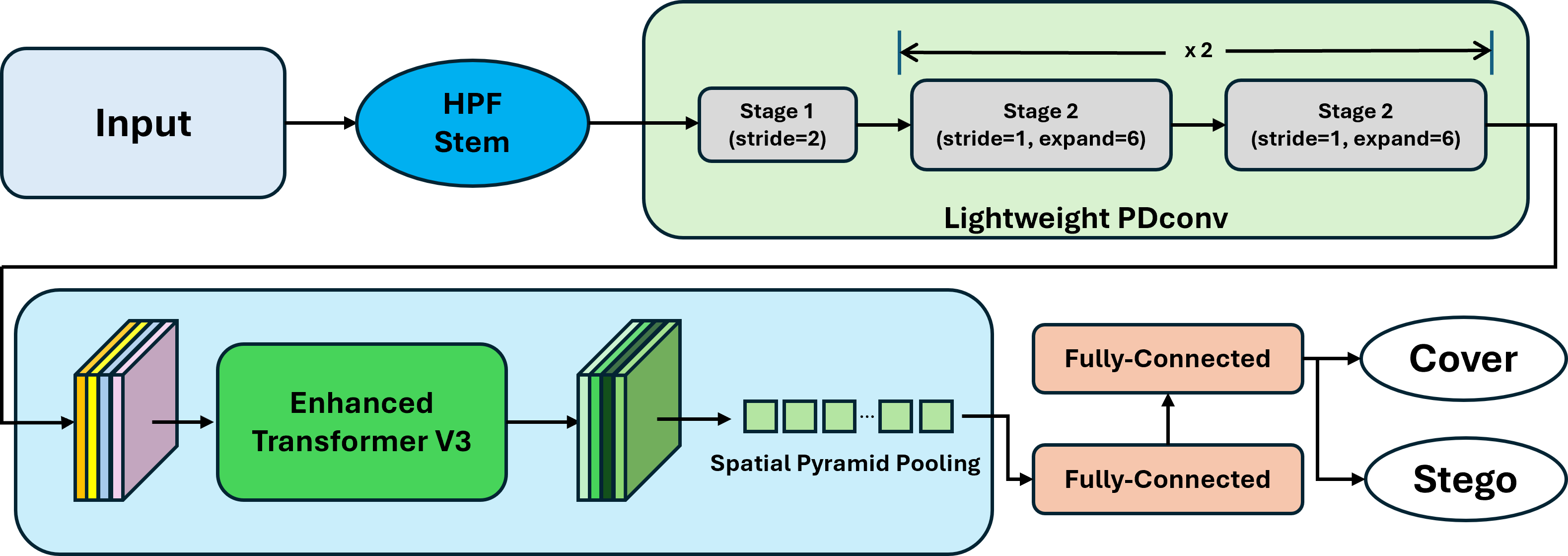}
  \caption{Overall architecture: proposed PENet+ is lightweight model.}
  \label{fig:arch}
\end{figure*}

\section{Proposed Method}
\label{sec:method}

\subsection{Efficiency Limitations in Existing Steganalysis Models}
\rev{Prior to the main discussion, we summarize key efficiency limitations in existing steganalysis models, as follows:}




\begin{itemize}
    \item \rev{\textbf{Fixed high-pass filter (HPF) bank without selection.}}
    \rev{Many CNN and Transformer steganalyzers use fixed SRM and Gabor residual filters as a front end, while redundancy and dataset dependent relevance are rarely analyzed, even though early residual extraction can dominate memory and computation~\cite{ye2017yenet,yedroudj2018cnn,srnet2018boroumand,penet2023}.}

    \item \rev{\textbf{Depthwise only pixel difference convolution (PDConv) inefficiency.}}
    \rev{PDConv improves residual sensitivity, but limited pointwise channel mixing can propagate wide feature maps through expensive high resolution spatial operations~\cite{penet2023,sandler2018mobilenetv2}.}

    \item \rev{\textbf{Full resolution multi-head self attention (MHSA) cost.}}
    \rev{Transformer based designs often preserve high resolution tokens, and MHSA cost grows quadratically with token count, making attention expensive for $512{\times}512$ inputs~\cite{dosovitskiy2021vit,liu2021swin,penet2023}.}

    \item \rev{\textbf{Oversized classifier bottlenecks.}}
    \rev{Large fully connected layers after spatial pyramid pooling (SPP) or global pooling can add nontrivial parameter and compute overhead, but this cost is often not analyzed in detail in prior work~\cite{he2015spp,li2018gcpnorm,deng2019gcpsteganalysis,zhang2020depthwise}.}
\end{itemize}

\smallskip
\noindent
Taken together, these design choices make existing steganalysis models computationally heavy and difficult to deploy in resource-constrained settings, highlighting the need for systematic efficiency-oriented refinement.
Based on these analyses, this section details an efficiency-oriented refinement of the original PENet~\cite{penet2023} into our lightweight and stability-oriented \textbf{PENet+}.
While PENet effectively combined pixel-difference convolution (PDConv) and transformer-based selective pooling, it suffered from heavy computational overhead due to dense attention and wide convolutional stacks. 
PENet+ revisits the HPF stem and convolutional backbone while preserving the original MHSA topology, aiming to maximize computational efficiency while maintaining steganalytic sensitivity under the following design improvements:
\begin{itemize}
\item \textbf{Activation–aware HPF selection.} A learnable HPF selector is trained on Y–channel residuals to produce per-filter importance scores, from which we select a balanced SRM–Gabor top-$K$ subset before training the main PENet+ model.
\item \textbf{MobileNetV2-inspired inverted residual backbone} replacing heavy stacks for lightweight spatial computation.
\item \textbf{Classifier bottleneck optimization} via reducing the FC1 input dimension after SPP, which accounts for most parameter/FLOPs savings with minimal accuracy loss.
\item \textbf{Activation–aware optimization} using PReLU for stable convergence under compression.
\end{itemize}
Table~\ref{tab:structure_penet_plus} summarizes the detailed architecture of the smallest variant, \textbf{PENet+--S}, while PENet+--M/L share the same layout but use wider backbone channels and FC1 inputs as listed in Table~\ref{tab:fc1_param}.

We consider a no-selection baseline and two selection modes built from the full SRM+Gabor bank (62 filters; 30 SRM, 32 Gabor). 
\textbf{v1} is the \emph{original PENet baseline} that uses all 62 filters (no selection). 
\rev{\textbf{v2} is an \emph{activation-aware pre-selection} strategy. Before training the main model, we run a short warm-up on the training set to measure per-filter activation energies and then fix a balanced top-$K$ subset.}
\textbf{v3} is a \emph{learnable activation-aware selector}: a lightweight gating network is trained on top of the fixed SRM+Gabor bank to predict per-filter importance scores, from which a balanced top-$K$ subset is selected; the final PENet+ model is then trained from scratch using only the selected filters.
Both v2 and v3 enforce SRM–Gabor balance using our scheduler in Algorithm~\ref{alg:hpf_balanced}. 
Unless specified, we adopt v3 as the default.

\begin{table*}[t]
\centering
\caption{Detailed structure configuration of the proposed \textbf{PENet+} framework for $512\times512$ color images in YCbCr space.
$K$ is the number of selected HPFs (here $K{=}31$), and $t$ is the expansion ratio of the inverted residual (IR) blocks. The configuration shown corresponds to the smallest variant, \textbf{PENet+--S} ($\widehat{C}{=}32$, $D_{\mathrm{in}}{=}672$).}
\label{tab:structure_penet_plus}
\setlength{\tabcolsep}{4.5pt}
\renewcommand{\arraystretch}{1.15}
\begin{tabular}{llcc}
\toprule
\textbf{Module} & \textbf{Name} &
\begin{tabular}{c}
\textbf{Input Kernel Size}\\
(width$\times$height)$\times$depth
\end{tabular} &
\begin{tabular}{c}
\textbf{Output Size}\\
(width$\times$height)$\times$channels
\end{tabular} \\
\midrule
\multirow{3}{*}{Stem Module} 
  & Input 
  & $(512\times512)\times3$ 
  & $(512\times512)\times3$ \\[1pt]
  & Separation \& HPF 
  & $(5\times5)\times31$ 
  & $(512\times512)\times31$ (per channel) \\[1pt]
  & Concatenation (Y/Cb/Cr) 
  & --- 
  & $(512\times512)\times93$ \\
\midrule
\multirow{5}{*}{Inverted Residual Backbone} 
  & Step 1 (IR, $t{=}1$) 
  & $(3\times3)\times93$, stride$=2$
  & $(256\times256)\times8$ \\[1pt]
\cmidrule(lr){2-4}
  & Step 2 (IR$\times2$, $t{=}6$) 
  & $(3\times3)\times6\times8$, stride$=2$ 
  & $(128\times128)\times16$ \\[1pt]
  & 
  & $(3\times3)\times6\times16$, stride$=1$ 
  & $(128\times128)\times16$ \\[1pt]
\cmidrule(lr){2-4}
  & Step 3 (IR$\times2$, $t{=}6$) 
  & $(3\times3)\times6\times16$, stride$=2$ 
  & $(64\times64)\times32$ \\[1pt]
  & 
  & $(3\times3)\times6\times32$, stride$=1$ 
  & $(64\times64)\times32$ \\
\midrule
\multirow{2}{*}{\shortstack{Enhanced Transformer Module\\ with Selective Pooling}}
  & MHSA block (K/V IR, $t{=}4$) 
  & $(3\times3)\times4\times32$
  & $(64\times64)\times32$ \\[1pt]
  & SPP
  & AvgPool $4\times4$, $2\times2$, $1\times1$ 
  & $1\times672$ (feature vector) \\
\midrule
\multirow{2}{*}{Classification Module} 
  & Fully-Connected 1 
  & 672  & 512 \\
  & Fully-Connected 2 
  & 512  & 2 \\
\bottomrule
\end{tabular}
\end{table*}

\subsection{Notation and Preliminaries}
Given a \textbf{YCbCr} image $\mathbf{I}\in\mathbb{R}^{3\times H\times W}$, each channel
$c\in\{Y,\mathrm{Cb},\mathrm{Cr}\}$ is filtered by a high-pass filter bank 
$\mathcal{F}=\{f_i\}_{i=1}^{62}$ containing 30 SRM and 32 Gabor filters 
(8 orientations, 2 scales, 2 phases). Each filter is set to $5{\times}5$ and convolved as:
\[
\mathbf{R}_c=\operatorname{TLU}\!\big(\,[f(\mathbf{I}_c)]_{f\in\mathcal{F}_{\mathrm{sel}}}\big), 
\qquad 
\mathbf{X}_0=\big[\mathbf{R}_Y;\,\mathbf{R}_{\mathrm{Cb}};\,\mathbf{R}_{\mathrm{Cr}}\big],
\]
where TLU clamps activations within $\pm\tau$ ($\tau{=}5.0$).
While PENet used all 62 filters, our model selects a compact, statistically meaningful subset
$\mathcal{F}_{\mathrm{sel}}$ ($K\!\ll\!62$) as described in Section~\ref{sec:hpf_selection}.
\noindent\textit{Channel note.}
Filter \emph{ranking} for pre-selection is computed on the Y channel only (Sec.~\ref{sec:hpf_selection}),
while the selected filters $\mathcal{F}_{\mathrm{sel}}$ are applied to all three YCbCr channels (Y, Cb, Cr) during training and inference.

\subsection{HPF Selection: Non-learning vs.\ Learning-based Selection}
\label{sec:hpf_selection}

We start from a 62-filter SRM+Gabor bank composed of 30 SRM and 32 Gabor filters, and construct a compact yet balanced HPF stem. All selection variants share a common balanced selector (Algorithm~\ref{alg:hpf_balanced}), which takes as input a score vector over the 62 filters and returns a subset of size $K$ while preserving the SRM/Gabor ratio and the orientation–phase coverage of the Gabor filters. By default, we set $K{=}31$, which yields a balanced split of $K_{\text{Gab}}{=}\lceil 31\cdot 32/62\rceil{=}16$ and $K_{\text{SRM}}{=}15$. Filter scoring (energy in v2, gate-based importance in v3) uses only the Y channel, whereas the final selected set $\mathcal{F}_{\mathrm{sel}}$ is applied to all YCbCr channels during both training and inference.
In this subsection, “Non-learning selection’’ refers to v2 (Algorithm~2), and “Learning-based selection’’ refers to v3 (Algorithm~3); both rely on the same balanced rule in Algorithm~\ref{alg:hpf_balanced}. 

The non-learning variant v2 is a rule-based scheme that uses activation statistics but introduces no additional trainable parameters. We first fix the 62-channel HPF stem and run a short warm-up pass on a subset of the training set (Y channel only) to compute the average absolute activation energy of each filter. The resulting 62-dimensional energy vector is then fed to Algorithm~\ref{alg:hpf_balanced} to obtain a balanced top-$K$ subset, which is kept fixed for the remainder of training. The overall activation-aware pre-selection process for v2 is summarized in Algorithm~2.

The learning-based variant v3 keeps the SRM+Gabor kernels themselves fixed, and instead learns channel importance via a lightweight selector network. We apply global average pooling (GAP) over the HPF stem outputs to obtain a 62-dimensional vector per sample, which is passed through a two-layer \rev{multi layer perceptron (MLP)} that produces per-filter gates in $(0,1)$ via a sigmoid. The selector is trained with a standard binary classification loss for steganalysis together with an $\ell_1$ sparsity penalty on the gates, encouraging filters that contribute more to the decision boundary to receive larger gate values. After pre-training, we average the gates over the training set to obtain a single importance score per filter and again invoke Algorithm~\ref{alg:hpf_balanced} under the same balance constraint to select the top-$K$ filters. We then instantiate \textit{PENet+} with an HPF stem that contains only the selected filters and train the full backbone from scratch. Although v3 requires a few additional epochs to train the selector, this stage remains lightweight (fixed HPF stem plus a small MLP), and the final selection step still reduces to sorting over the same 62 filters once. The overall learning-based selection process for v3 is summarized in Algorithm~3.

The original PENet paper also reports a \emph{Learnable HPF} setting, where the SRM+Gabor kernels are treated as trainable parameters~\cite{penet2023}. However, for both $256{\times}256$ and $512{\times}512$ inputs, this configuration is reported to suffer from unstable convergence, with detection accuracy remaining around 50\% (i.e., close to random guessing). In contrast, \textit{PENet+} never updates HPF weights directly; instead, it only learns which fixed SRM+Gabor filters to keep, which in practice yields more stable convergence and a better accuracy–efficiency trade-off.

The HPF budget $K$ is chosen by sweeping $K\in\{15,31,48\}$ while keeping the SRM/Gabor ratio and the backbone/attention topology fixed. Across activation functions and classifier widths ($D_{\text{in}}\in\{672,1344,2688\}$), $K{=}31$ either matches or slightly outperforms $K{=}48$ while using fewer HPF channels and less computation. In contrast, $K{=}15$ tends to underfit because its orientation and phase coverage is too limited. We therefore adopt $K{=}31$ (16 Gabor and 15 SRM filters) as the default HPF budget and use other values of $K$ only in ablation studies (Sec.~\ref{sec:ablation-results}).

\begin{algorithm}[t]
\caption{Balanced SRM--Gabor Top-$K$ Selector}
\label{alg:hpf_balanced}
\begin{algorithmic}[1]
\REQUIRE Full bank $\mathcal{F}=\mathcal{F}_{\text{SRM}}\cup\mathcal{F}_{\text{Gab}}$, target size $K$, 
optional energies $\{E_i\}$ (higher is better; if unavailable, fall back to a deterministic order)
\STATE Set $K_{\text{Gab}}\!\leftarrow\!\lceil K\cdot 32/62\rceil$, \; $K_{\text{SRM}}\!\leftarrow\!K-K_{\text{Gab}}$
\STATE \textbf{SRM selection:} 
\STATE \hspace{0.8em}If energies $\{E_i\}$ exist: take top-$K_{\text{SRM}}$ from $\mathcal{F}_{\text{SRM}}$ by $E_i$. 
Else: take the first $K_{\text{SRM}}$ from a fixed canonical SRM list (edge/2nd-derivative, cross, diagonals).
\STATE \textbf{Gabor selection (diversity-aware):}
\STATE \hspace{0.8em}Partition $\mathcal{F}_{\text{Gab}}$ into pools by orientation $\theta\in\{0^\circ,22.5^\circ,\dots,157.5^\circ\}$ and phase $\phi\in\{0,\pi/2\}$.
\STATE \hspace{0.8em}Within each $(\theta,\phi)$ pool, order filters by key: 
\STATE \hspace{1.6em}\textbullet\;\;If $\{E_i\}$ exist: descending $E_i$; 
\STATE \hspace{1.6em}\textbullet\;\;Else: increasing spatial frequency (low$\rightarrow$high).
\STATE \hspace{0.8em}Round-robin over orientations; at each step, alternate phase $\phi$ and pop the current pool’s top until $K_{\text{Gab}}$ are chosen (skip empty pools).
\RETURN $\mathcal{F}_{\text{sel}}=\mathcal{F}_{\text{Gab}}^{(K_{\text{Gab}})}\cup\mathcal{F}_{\text{SRM}}^{(K_{\text{SRM}})}$
\end{algorithmic}
\end{algorithm}

\begin{algorithm}[t]
\caption{Activation-aware HPF Pre-selection (v2)}
\label{alg:hpf_pre}
\begin{algorithmic}[1]
\REQUIRE Training subset $\mathcal{D}_{\text{warmup}}$, full bank $\mathcal{F}$, target size $K$
\STATE For all $f_i\in\mathcal{F}$, initialize $E_i\!\leftarrow\!0$
\FORALL{$I\in\mathcal{D}_{\text{warmup}}$} \hfill \textit{// no augmentation}
  \STATE Convert to YCbCr and take $I_Y$
  \FORALL{$f_i\in\mathcal{F}$}
    \STATE $A\!\leftarrow\!f_i(I_Y)$ \hfill \textit{// same padding $5\!\times\!5$}
    \STATE $E_i\!\leftarrow\!E_i+\|A\|_1$
  \ENDFOR
\ENDFOR
\STATE Normalize $E_i\!\leftarrow\!E_i/|\mathcal{D}_{\text{warmup}}|$
\STATE $\mathcal{F}_{\text{sel}}\!\leftarrow\!\textsc{BalancedSelector}(\mathcal{F},K,\{E_i\})$ \hfill \textit{// Alg.~\ref{alg:hpf_balanced}}
\STATE Freeze $\mathcal{F}_{\text{sel}}$ and train the model with this stem
\end{algorithmic}
\end{algorithm}

\begin{algorithm}[t]
\caption{Learnable HPF Selector (v3; default)}
\label{alg:hpf_post}
\begin{algorithmic}[1]
\REQUIRE Training set $\mathcal{D}$, full bank $\mathcal{F}$, target size $K$, selector training budget $T_{\text{sel}}$
\STATE Initialize selector network $S_\theta$ (two-layer MLP) that maps globally pooled HPF responses to per-filter gates.
\STATE Freeze SRM+Gabor kernels in $\mathcal{F}$.
\FOR{$t = 1$ to $T_{\text{sel}}$}
  \STATE Sample mini-batch $I\subset\mathcal{D}$ and convert each image to YCbCr.
  \STATE Apply $\mathcal{F}$ to the Y channel and globally average pool each filter response to obtain feature vector $\mathbf{r}(I)$.
  \STATE Compute gates $\mathbf{g}(I)=S_\theta(\mathbf{r}(I))$ and steganalysis logits.
  \STATE Compute classification loss $L_{\text{cls}}$ and sparsity loss $L_{\text{sp}}=\lambda_{\text{sp}}\|\mathbf{g}(I)\|_1$.
  \STATE Set total loss $L=L_{\text{cls}}+L_{\text{sp}}$ and update $\theta$ via back-propagation.
\ENDFOR
\STATE Estimate per-filter importance scores $E_i$ by averaging gates $g_i(I)$ over $\mathcal{D}$.
\STATE $\mathcal{F}_{\text{sel}}\!\leftarrow\!\textsc{BalancedSelector}(\mathcal{F},K,\{E_i\})$ \hfill \textit{// Alg.~\ref{alg:hpf_balanced}}
\STATE Instantiate the final PENet+ model with HPF stem $\mathcal{F}_{\text{sel}}$ and train it from scratch.
\end{algorithmic}
\end{algorithm}

\begin{figure*}[!t]
  \centering
  \includegraphics[width=0.95\linewidth]{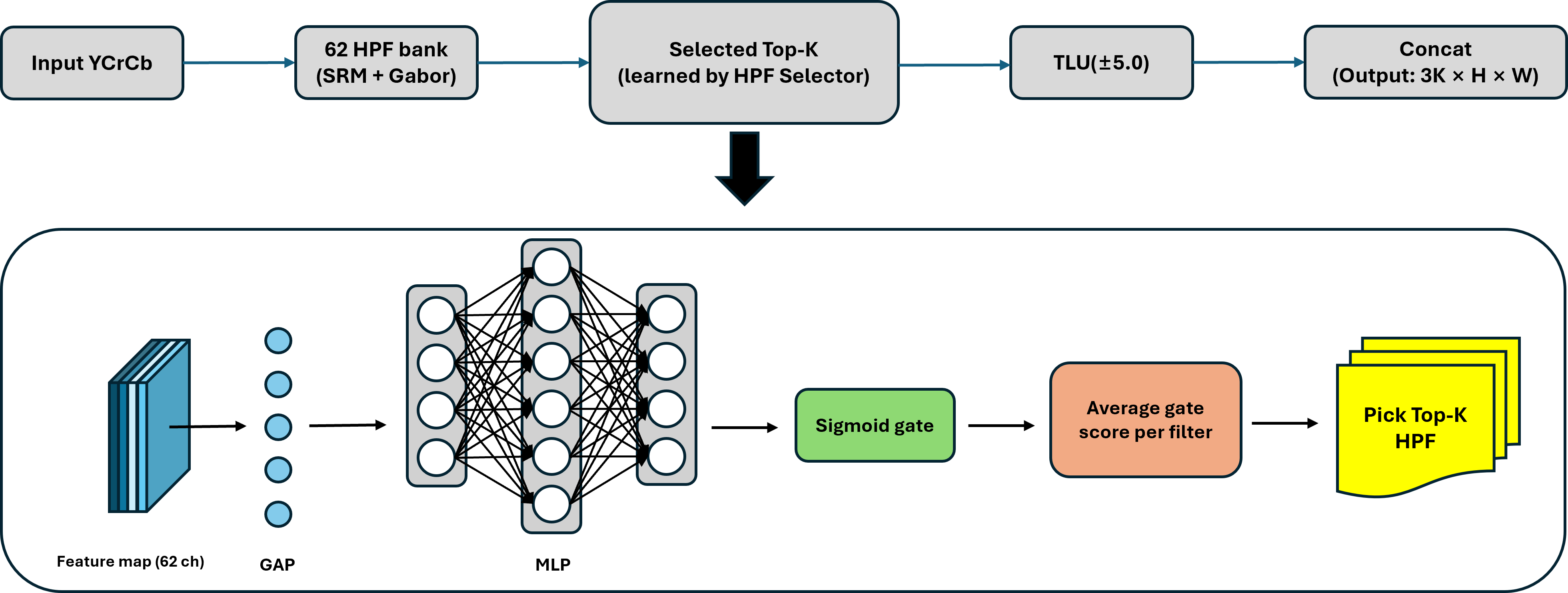}
  \caption{HPF stem with activation-aware Top-$K$ selection (balanced SRM–Gabor), followed by TLU ($\pm 5.0$) and channel concatenation.}
  \label{fig:hpf_selector}
\end{figure*}

\subsection{Lightweight Backbone via Inverted Residuals}
\label{sec:backbone}

PDConv stacks in PENet are effective at capturing micro-texture artifacts, but they are computationally heavy because they include redundant convolutional paths. They also lack pointwise projection layers and rely only on depthwise-like operations, which is unfavorable for mobile and edge deployment.
To improve efficiency, PENet+ replaces the PDConv stages with MobileNetV2-style inverted residual (IR) blocks, consisting of $1{\times}1~\text{expand}
~\rightarrow~3{\times}3~\text{depthwise (stride)}
~\rightarrow~1{\times}1$ linear projection.
Each IR block follows the expand–depthwise–project paradigm~\cite{sandler2018mobilenetv2}. Given input
$\mathbf{X}\!\in\!\mathbb{R}^{C_{\mathrm{in}}\times H\times W}$, the block first expands the channels by a factor $r$ with a $1{\times}1$ convolution, then applies a $3{\times}3$ depthwise convolution with stride $s\!\in\!\{1,2\}$, and finally uses a linear $1{\times}1$ projection to produce $C_{\mathrm{out}}$ channels. Batch normalization (BN) follows each convolution and PReLU is used as the default activation $\phi(\cdot)$. When $s{=}1$ and $C_{\mathrm{in}}{=}C_{\mathrm{out}}$, a residual shortcut is added. The last $1{\times}1$ layer has no nonlinearity, which preserves information through a linear bottleneck~\cite{sandler2018mobilenetv2}:
\begin{IEEEeqnarray}{rCl}
\mathbf{U} & = & \mathrm{BN}\!\big(\mathrm{Conv}_{1\times 1}(\mathbf{X};\, C_{\mathrm{exp}}{=}rC_{\mathrm{in}})\big), \\
\mathbf{V} & = & \mathrm{BN}\!\big(\mathrm{DWConv}_{3\times 3}^{(s)}(\phi(\mathbf{U}))\big), \\
\mathbf{Y} & = & \mathrm{BN}\!\big(\mathrm{Conv}_{1\times 1}(\phi(\mathbf{V});\, C_{\mathrm{out}})\big). \IEEEyesnumber
\end{IEEEeqnarray}

\begin{figure}[t]
\centering
\includegraphics[width=0.9\linewidth]{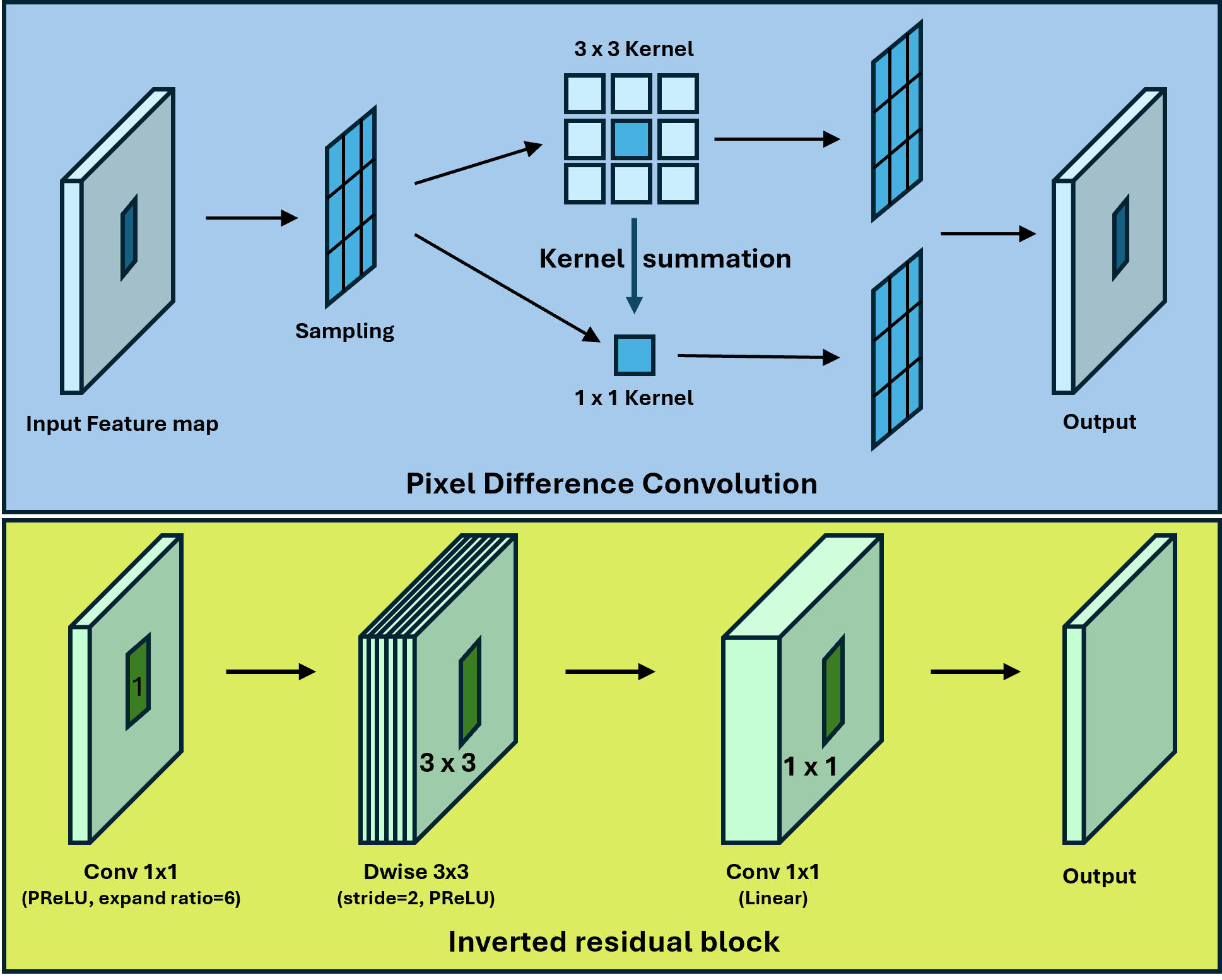}
\caption{Comparison between the original PDConv-based backbone and the proposed inverted residual (IR) backbone. The IR design replaces heavy full convolutions with expand–depthwise–project blocks, reducing spatial FLOPs while preserving residual sensitivity.}
\label{fig:pdconv_vs_ir}
\end{figure}

Compared with PDConv stacks, which extract pixel-difference cues but keep wide feature maps flowing through expensive spatial convolutions, IR blocks shift most channel mixing to $1{\times}1$ pointwise layers and use cheap depthwise spatial filtering. This design yields large savings at high resolutions while preserving representational capacity through the expansion factor $r$. A schematic comparison between the original PDConv stack and the proposed IR block is shown in Fig.~\ref{fig:pdconv_vs_ir}.

For an IR block with spatial size $(H,W)$, expansion $r$, and output channels $C_{\mathrm{out}}$, the FLOPs are
\begin{IEEEeqnarray}{rCl}
\mathrm{FLOPs}_{1{\times}1}^{\text{exp}} & = & HW\, C_{\mathrm{in}}\,(rC_{\mathrm{in}}), \\
\mathrm{FLOPs}_{\text{DW}} & = & HW \cdot 9 \,(rC_{\mathrm{in}}), \\
\mathrm{FLOPs}_{1{\times}1}^{\text{proj}} & = & HW \,(rC_{\mathrm{in}})\, C_{\mathrm{out}}. \IEEEyesnumber
\end{IEEEeqnarray}
Replacing a full $3{\times}3$ convolution with $9HWC^{2}$ FLOPs by a depthwise $3{\times}3$ convolution with $9HWC$ FLOPs followed by two $1{\times}1$ pointwise projections with $2HWC^{2}$ FLOPs reduces the spatial cost by roughly a factor of $C$ at high resolutions, while maintaining channel mixing capacity in the $1{\times}1$ layers.

In PENet+, we set $r{=}6$ in the early stages and use stride $s{=}2$ at stage boundaries to perform downsampling. All convolutions are followed by BN, and PReLU is used everywhere except after the final linear bottleneck. Squeeze-and-excitation and DropPath are omitted to keep the critical path short on edge devices. 
This backbone configuration maintains sensitivity to residual artifacts while substantially reducing feature-map computation, with an empirical reduction of about $60\%$ FLOPs in the high-resolution stages.

\begin{figure*}[!t]
  \centering
  \includegraphics[width=0.95\linewidth]{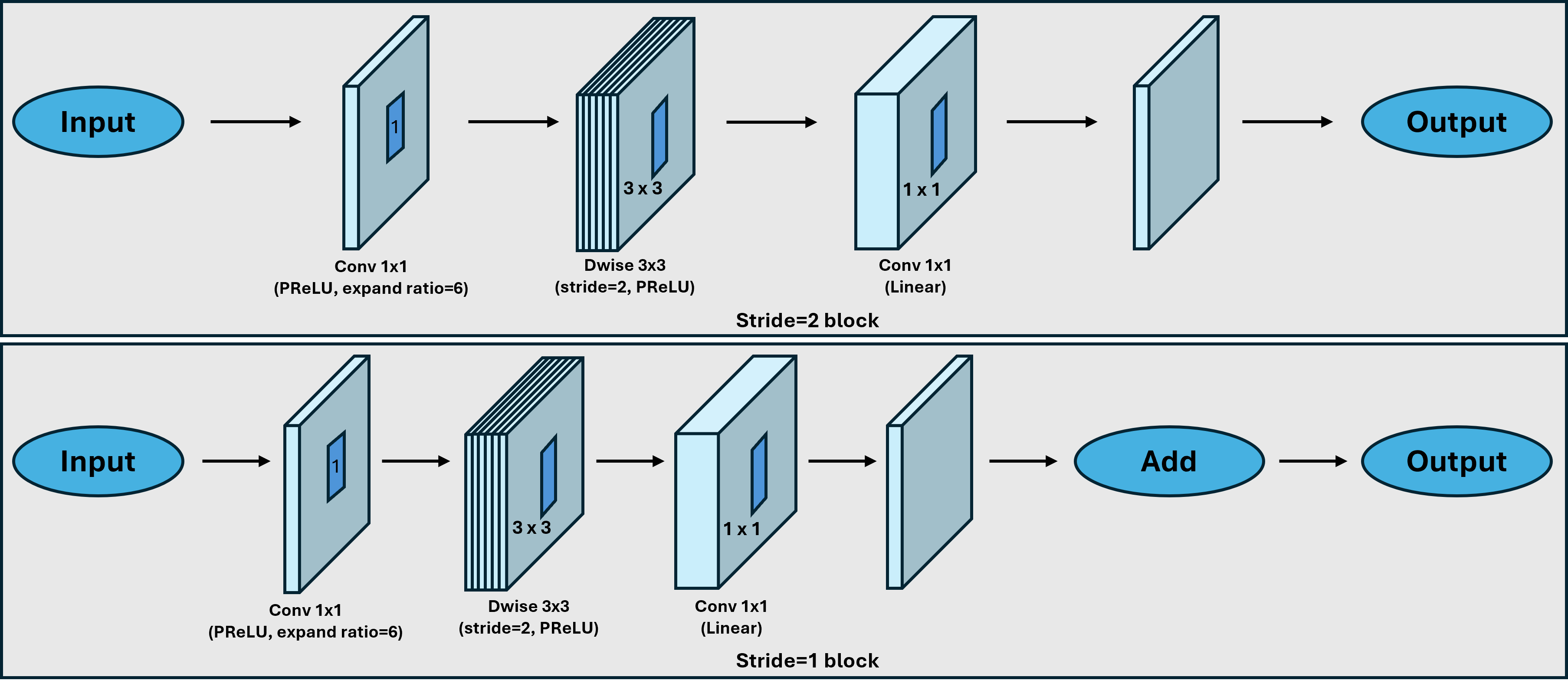}
  \caption{Lightweight backbone based on inverted residual (IR) blocks with stage-wise stride/expansion and channel widths.}
  \label{fig:ir_backbone}
\end{figure*}

\begin{table*}[t]
\centering
\caption{Design-by-design comparison between PENet and the proposed PENet+.}
\label{tab:compare}
\begin{tabularx}{\textwidth}{l X X}
\toprule
\textbf{Aspect} & \textbf{PENet (baseline)} & \textbf{PENet+ (ours)} \\
\midrule
HPF stem &
62 fixed SRM+Gabor filters applied to RGB & Activation–aware subset ($K\ll 62$), balanced SRM–Gabor, selected via activation-energy (v2/v3) \\
\midrule
Early feature extractor & PDConv stacks (pixel-difference depthwise filters) & MobileNetV2 inverted residual blocks for lightweight spatial processing \\
\midrule
Attention module &
EnhanceMHSA with PDConv-based depthwise K/V branches (stride $=2$) and learned relative bias $B$ & Same EnhanceMHSA topology (Q from layer-normalized tokens, K/V from conv branches with learned bias $B$); in PENet+ the K/V branches are implemented with lightweight IR-based conv blocks and narrower channels, reducing attention FLOPs without changing the token layout \\
\midrule
Activation & ReLU (fixed slope) & PReLU (channel-wise learnable slope); preserves negative residual cues important for steganalysis \\
\bottomrule
\end{tabularx}
\end{table*}

\subsection{Classifier Streamlining (FC1)}
\label{sec:fc}

In PENet, the first fully connected layer (FC1) after spatial pyramid pooling (SPP) accounts for a notable portion of the total parameters. Let $X \in \mathbb{R}^{C\times H\times W}$ be the feature map before SPP and let
\[
S = \sum_{p\in\{4,2,1\}} p^2 = 16 + 4 + 1 = 21
\]
be the total number of pooled cells. The SPP output is then a vector of dimension
\[
D_{\mathrm{in}} = C \cdot S,
\]
which is fed into FC1 of size $D_{\mathrm{hid}}$ (bias omitted), yielding $D_{\mathrm{in}} \times D_{\mathrm{hid}}$ parameters.

In the original PENet, the channel width before SPP is fixed to $C = 256$, so the classifier input becomes
\[
D_{\mathrm{in}}^{\text{base}} = 256 \times 21 = 5376,
\]
and FC1 has $5376 \times 1024$ parameters, which is about $5.5$M. This makes FC1 one of the dominant contributors to model size and FLOPs.

PENet+ reduces this cost by explicitly designing a narrower backbone width $\widehat{C}$ before SPP, with $\widehat{C} \in \{32,64,128\}$. The SPP–FC1 interface becomes
\[
D_{\mathrm{in}} = \widehat{C}\cdot S.
\]
For example, $\widehat{C}=128$ yields $D_{\mathrm{in}}=2688$ (about $2.8$M parameters for FC1 with $D_{\mathrm{hid}}=1024$), $\widehat{C}=64$ yields $D_{\mathrm{in}}=1344$, and $\widehat{C}=32$ combined with a smaller hidden size $D_{\mathrm{hid}}=512$ reduces FC1 to $672 \times 512$ parameters (about $0.34$M). Since $D_{\mathrm{in}}$ scales linearly with $\widehat{C}$, the FC1 parameter count and FLOPs also scale linearly. In our experiments, this streamlining reduces classifier parameters by roughly a factor of $4$ to $16$ while keeping detection accuracy. The self-attention topology from PENet is kept unchanged; most efficiency gains come from the lightweight backbone and HPF selection, with classifier narrowing serving as a complementary reduction of the SPP–FC1 interface.

\begin{table}[t]
\centering
\caption{FC1 input dimension reduction with different backbone channel.}
\label{tab:fc1_param}
\begin{tabular}{lrrr}
\toprule
model & $\widehat{C}$ & $D_{\mathrm{in}}$ & $D_{\mathrm{hid}}$ \\
\midrule
Baseline     & 256 & 5376 & 1024 \\
PENet+--L    & 128 & 2688 & 1024 \\
PENet+--M    &  64 & 1344 & 1024 \\
PENet+--S    &  32 &  672 &  512 \\
\bottomrule
\end{tabular}
\end{table}

\subsection{Activation Function Optimization}
\label{sec:act}
Activation functions critically influence gradient propagation in residual architectures.  
We tested ReLU, ReLU6, LeakyReLU, SiLU, and PReLU under identical training settings.
We note that {PReLU outperformed all variants}, consistently achieving faster convergence and higher accuracy.
Unlike ReLU6, which clips negative activations and causes information loss, PReLU preserves weak negative responses through channel-wise learned slopes $\alpha_c$.  
This enables adaptive nonlinearity per feature map and improves learning under varying JPEG quality factors and payload sizes.  
Empirically, PReLU consistently showed the best overall performance and more stable convergence than the other activation functions.

\subsection{Optimization and Implementation}
We adopt AdamW with decoupled weight decay; normalization and PReLU parameters are exempt from decay.  
Cosine annealing is used for learning rate scheduling.  
Before evaluation, BatchNorm statistics are recalibrated once using training data (no gradients).  
Augmentations include $90^\circ$ rotations and horizontal flips.
Default hyperparameters: $K{=}31$, $s{=}2$, $h{=}2$, $d_k{=}d_v{=}64$, TLU $\tau{=}5.0$, SPP $\{4,2,1\}$.

\subsection{Attention (Topology Preserved with Lightweight K/V Branches)}
\label{sec:mhsa}
We keep the multi-head self attention (MHSA) topology of PENet---the scaled dot-product formulation, residual connection, and token layout on the query side---while replacing the convolutional $K/V$ branches with lightweight inverted residual (IR) blocks.

Let $x\in\mathbb{R}^{B\times C\times H\times W}$ be the input feature map before attention. We first form the \emph{query} token sequence $X_q\in\mathbb{R}^{B\times T_q\times C}$ by flattening spatial positions ($T_q = H W$) and applying layer normalization:
\[X_q = \mathrm{LN}\!\big(\mathrm{reshape}(x)\big),\]
and then compute
\[Q = X_q W_Q \in \mathbb{R}^{B\times T_q\times (h d_k)},\]
where $h$ is the number of heads and $d_k$ is the key dimension per head.
The \emph{key} and \emph{value} sequences are obtained from the IR-based $K/V$ backbones. We apply two shared-structure inverted residual stacks, $\phi_K(\cdot)$ and $\phi_V(\cdot)$, each operating on $x$ with stride $s$ (and expansion ratio $t$), which produce downsampled feature maps of spatial size $H'\times W'$:
\[F_K = \phi_K(x),\quad F_V = \phi_V(x),\quad H' = H/s,\; W' = W/s.\]
After flattening, we obtain
$X_K,X_V\in\mathbb{R}^{B\times T_k\times C}$ with $T_k = H' W'$, and \[K = X_K W_K,\quad V = X_V W_V,\]
where $W_K,W_V$ are learned linear projections.
For each head, we reshape $Q,K,V$ into $\mathbb{R}^{B\times h\times T_q\times d_k}$ and $\mathbb{R}^{B\times h\times T_k\times d_k}$, and compute scaled dot-product attention with a learned bias tensor $B\in\mathbb{R}^{1\times h\times T_q\times T_k}$:
\[A = \operatorname{softmax}\!\left(\frac{QK^\top}{\sqrt{d_k}} + B\right),\]
\[O = A V,\quad Y = \mathrm{Proj}(O),\]
where $\mathrm{Proj}$ is the output projection that maps the concatenated heads back to $C$ channels and $Y$ is reshaped to $\mathbb{R}^{B\times C\times H\times W}$.
Finally, a residual connection is added:
\[Z = Y + x.\]
In summary, PENet+ preserves the original MHSA computation on the query side and the residual topology of PENet, but replaces the PDConv-based $K/V$ backbones with lightweight IR blocks and narrower channels. This reduces attention FLOPs while leaving the attention mechanism and tokenization scheme on the query side unchanged.

\begin{figure*}[!t]
  \centering
  \includegraphics[width=0.95\linewidth]{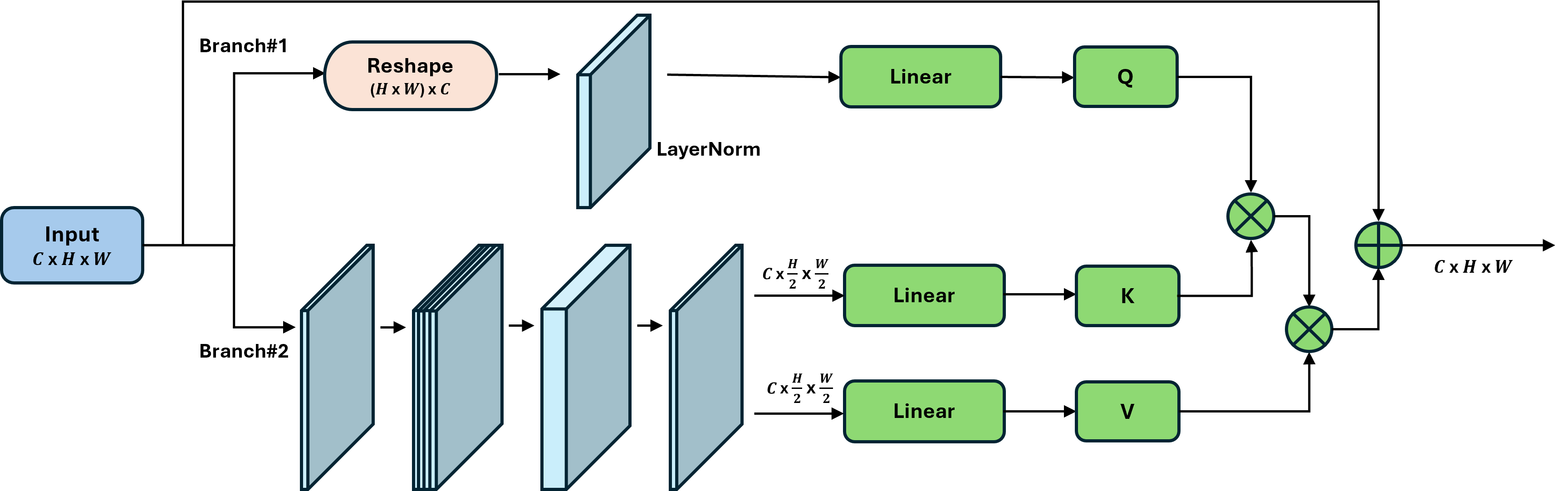}
  \caption{Transformer head (topology unchanged from PENet). Projections use $d_k{=}d_v{=}64$ (2 heads), followed by standard MHSA and reshape back to $32\times64\times64$.}
  \label{fig:trans_v3}
\end{figure*}

\subsection{Activation-Energy Visualization}
\label{sec:hpf_activation}

To inspect where the network attends without relying on class-targeted CAMs, we visualize a \emph{class-agnostic per-pixel activation energy} from selected layers.
Given a feature tensor $F \in \mathbb{R}^{C \times H \times W}$, we compute
\[
E(x,y) = \frac{1}{C}\sum_{c=1}^{C} \bigl|F_c(x,y)\bigr|,
\]
min–max normalize $E$ to $[0,1]$, resize it to the original image resolution, and overlay a colormap on the RGB image with a fixed opacity.
This map highlights spatial regions where a layer is strongly activated, independently of the predicted class.
We probe five stages spanning early residual extraction to deeper semantic processing:
(i) the HPF+TLU output immediately after the stem, and (ii–v) four successive groups in the backbone.
Earlier stages emphasize SRM/Gabor-like high-frequency residuals, whereas deeper stages reflect progressively higher-level patterns produced by PD/IR blocks.
Because the visualization does not use class gradients, it is complementary to class-discriminative CAMs and serves as a proxy for residual saliency preserved by our HPF selection and lightweight backbone.

\section{Experiments and Discussion}
\subsection{Experimental Setup}
\label{sec:exp-setup}


\rev{
We start from a pool of 24{,}000 ALASKA2 JPEG QF90 color images. We randomly sample 5{,}000 cover images using a fixed random seed (Seed = 2025) to construct matched cover stego pairs, and split the selected covers into 4{,}000 for training, 500 for validation, and 500 for internal testing. 
For each cover image in each split, we generate the corresponding stego images using nsF5~\cite{filler2007nsf5}, J-UNIWARD~\cite{holub2014junward}, and UERD~\cite{guo2015uerd} at payloads 0.2 and 0.4 bits per pixel (bpc).
The final evaluation set is built from the remaining 19{,}000 cover images that are not used in training, validation, internal testing, or selector warmup, together with their corresponding stego counterparts generated under the same embedding settings.
We explicitly ensure \textbf{no overlap between training and evaluation}: the training, validation, internal test, and evaluation sets are mutually disjoint with no shared images.
}

\rev{
\noindent\textbf{Statistical reporting.}
We repeat key results under the representative activation setting (PReLU) for $N{=}3$ independent runs with different training seeds and report mean and standard deviation (e.g., Table~\ref{tab:prelu_meanstd_512}).
All runs use the same fixed dataset split and preprocessing.
}

We report per-image accuracy, the number of parameters (Params) and FLOPs as the main indicators of detection performance and computational efficiency.
\rev{We additionally evaluate a $256{\times}256$ setting by extracting a center crop of size $256{\times}256$ from each image, using the same split protocol and training recipe.}



\subsection{Quantitative Evaluation}
Before presenting our results on ALASKA2 QF90, we briefly recap the original PENet setting in \cite{penet2023}. PENet was evaluated on an ALASKA2 $512{\times}512$ TIFF color subset using CMD-C-HILL~\cite{li2014hill} at 0.4~bpc, where PENet achieved 80.03\% detection accuracy and outperformed SID~\cite{tsang2018sid} (75.82\%) and SiaStegNet~\cite{you2021siastegnet} (50.45\%).
In contrast, our experiments (Table~\ref{tab:acc_flops_full}) use JPEG-compressed ALASKA2 QF90 images at $512{\times}512$~\cite{cogranne2019alaska,butora2021pretrain,yousfi2019breakingalaska,yousfi2020imagenetjpeg,yousfi2021improvingeffnet} and evaluate nsF5, J-UNIWARD, and UERD at 0.2 and 0.4~bpc. All PENet+ variants are trained under identical preprocessing and optimization conditions. We compare per-image detection accuracy together with model complexity in terms of parameters and FLOPs.

\begin{table*}[t]
\centering
\caption{Accuracy and FLOPs comparison on ALASKA2 QF90 (512$\times$512).
The first row is our re-trained and re-evaluated \textbf{PENet} baseline under the same ALASKA2 JPEG QF90 protocol, and the remaining rows are \textbf{PENet+} variants with HPF=31 while sweeping $D_{\text{in}}\in\{672,1344,2688\}$.}
\label{tab:acc_flops_full}
\small
\setlength{\tabcolsep}{6pt}
\renewcommand{\arraystretch}{1.2}
\begin{tabularx}{\textwidth}{l c c c *{6}{>{\centering\arraybackslash}X}}
\toprule
 & & & &
\multicolumn{2}{c}{\textbf{nsF5}} &
\multicolumn{2}{c}{\textbf{J-UNIWARD}} &
\multicolumn{2}{c}{\textbf{UERD}} \\
\cmidrule(lr){5-6}\cmidrule(lr){7-8}\cmidrule(lr){9-10}
\textbf{Activation} &
$\mathbf{D_{\text{in}}}$ &
\textbf{Params (M)} &
\textbf{FLOPs (G)} &
\textbf{0.2 bpc} & \textbf{0.4 bpc} &
\textbf{0.2 bpc} & \textbf{0.4 bpc} &
\textbf{0.2 bpc} & \textbf{0.4 bpc} \\
\midrule
\textbf{ReLU} & 5376 & 16.16 & 67.5
&0.7447 & 0.9311
&0.5701 & 0.7750
&0.5750 & 0.7757
\\
\midrule
\multirow{3}{*}{\textbf{ReLU6}}
& 672  &  8.80 & 2.12
&0.7338 & 0.8976
&0.7544 & 0.8917
&0.7187 & 0.8790
\\
& 1344 &  9.98 & 2.93
&0.7712 & 0.9294 
&0.7832 & 0.8953 
&0.7541 & 0.8807 
\\
& 2688 & 11.86 & 5.85
&0.7813  & \textbf{0.9429} 
&0.7178  & 0.8628 
&0.7349  & 0.8633 
\\
\midrule
\multirow{3}{*}{\textbf{LeakyReLU}}
& 672  &  8.80 & 2.12
&0.7315 & 0.9180 
&0.7983 & 0.9040 
&0.7334 & 0.8728 
\\
& 1344 &  9.98 & 2.93
&0.7665 & 0.9174
&0.8329 & 0.8800
&0.7215 & 0.8726
\\
& 2688 & 11.86 & 5.85
&0.7483 & \textbf{0.9439}
&0.6824 & 0.8503
&0.7239 & 0.8801
\\
\midrule
\multirow{3}{*}{\textbf{PReLU}}
& 672  &  8.80 & 2.12
&0.7649 & 0.9070 
&0.8053 & 0.9040
&0.7260 & 0.8613 
\\
& 1344 &  9.98 & 2.93
&0.7601 & 0.9235 
&0.8472 & 0.9356
&0.7968 & 0.8850 
\\
& 2688 & 11.86 & 5.85
&0.7914 & \textbf{0.9481}
&0.7104 & 0.8552 
&0.7308 & 0.8849
\\
\midrule
\multirow{3}{*}{\textbf{SiLU (Swish)}}
& 672  &  8.80 & 2.12
&0.7347 & 0.9048 
&0.7965 & 0.9106
&0.6792 & 0.8647
\\
& 1344 &  9.98 & 2.93
&0.7545 & 0.9324
&0.7097 & 0.9081
&0.7248 & 0.8752
\\
& 2688 & 11.86 & 5.85
&0.7708 & \textbf{0.9415} 
&0.6930 & 0.8600
&0.6991 & 0.8724
\\
\bottomrule
\end{tabularx}
\end{table*}

\begin{revblock}
We additionally report mean and standard deviation over $N{=}3$ runs for the key $512{\times}512$ PReLU setting in Table~\ref{tab:prelu_meanstd_512}.
\end{revblock}

\newcommand{\mstd}[2]{\begin{tabular}[c]{@{}c@{}}#1\\[-1pt]{\scriptsize(#2)}\end{tabular}}

\begin{table}[t]
\centering
\caption{\rev{Results on ALASKA2 QF90 at $512{\times}512$ with PReLU at 0.4 bpc. Accuracy is reported as mean(std) over $N{=}3$ runs.}}
\label{tab:prelu_meanstd_512}
\scriptsize
\setlength{\tabcolsep}{4pt}
\renewcommand{\arraystretch}{1.15}

\begin{tabularx}{\columnwidth}{>{\centering\arraybackslash}m{3.0cm} *{3}{>{\centering\arraybackslash}X}}
\toprule
& \multicolumn{3}{c}{\rev{$\mathbf{D_{\text{in}}}$}} \\
\cmidrule(lr){2-4}
\rev{\textbf{Embedding algorithm}} &
\rev{\textbf{672}} & \rev{\textbf{1344}} & \rev{\textbf{2688}} \\
\midrule

\rev{\textbf{nsF5 (0.4 bpc)}} &
\rev{0.9034(0.0029)} & \rev{0.9260(0.0056)} & \rev{0.9454(0.0089)} \\

\rev{\textbf{J-UNIWARD (0.4 bpc)}} &
\rev{0.8918(0.0089)} & \rev{0.9197(0.0113)} & \rev{0.8662(0.0078)} \\

\rev{\textbf{UERD (0.4 bpc)}} &
\rev{0.8561(0.0037)} & \rev{0.8795(0.0046)} & \rev{0.8886(0.0027)} \\

\bottomrule
\end{tabularx}
\end{table}

\begin{table}[t]
\centering
\caption{HPF count sweep (best train accuracy per activation).
Numbers are the best over FC1 inputs $D_{\text{in}}\in\{672,1344,2688\}$.}
\label{tab:k_sweep}
\begin{tabular}{lcccc}
\toprule
Algorithm & Act & $K{=}15$ & $K{=}31$ & $K{=}48$ \\
\midrule
nsF5@0.4      & PReLU & 0.9214 & \textbf{0.9481} & 0.9402 \\
nsF5@0.4      & SiLU  & 0.9362 & \textbf{0.9415} & 0.9409 \\
J-UNIWARD@0.4 & PReLU & 0.9085 & \textbf{0.9356} & 0.9115 \\
J-UNIWARD@0.4 & SiLU  & 0.8541 & \textbf{0.9106} & 0.9038 \\
UERD@0.4      & PReLU & 0.8827 & \textbf{0.8850} & 0.8653 \\
UERD@0.4      & SiLU  & 0.8567 & 0.8752 & \textbf{0.8824} \\
\bottomrule
\end{tabular}
\end{table}

\begin{revblock}
\subsubsection{Results at $256{\times}256$}
We additionally evaluate at $256{\times}256$ to assess scalability across resolutions.
For this setting, inputs are obtained via a deterministic center crop from the decoded $512{\times}512$ images (no resizing).
Therefore, the $256{\times}256$ experiment should be regarded as a controlled resolution stress test under the same original image pool and split protocol, rather than as a separate benchmark on native $256{\times}256$ ALASKA2 images.
Table~\ref{tab:acc_flops_256} reports detection performance together with Params and FLOPs at $256{\times}256$, and we report PENet+ under the representative activation setting (PReLU) consistent with the $512{\times}512$ results.
\end{revblock}

\begin{table}[t]
\centering
\caption{Comparison between activation-based HPF selection and random selection
(5-run average) for nsF5@0.4 with PReLU.}
\label{tab:k_random}
\setlength{\tabcolsep}{6pt}
\renewcommand{\arraystretch}{1.1}
\begin{tabular}{c l c c}
\toprule
$K$ & Selection method & Accuracy & $\Delta$ Accuracy \\
\midrule
\multirow{2}{*}{15} 
& Random & 0.9092 & --        \\
& Activation-based (ours) & 0.9214 & +0.0122   \\
\midrule
\multirow{2}{*}{31} 
& Random & 0.9319 & --        \\
& Activation-based (ours) & 0.9481 & +0.0162    \\
\bottomrule
\end{tabular}
\end{table}

\begin{table*}[t]
\centering
\caption{\rev{Accuracy and FLOPs comparison on ALASKA2 QF90 at $256{\times}256$.
Inputs are obtained via deterministic center crops from decoded $512{\times}512$ images (no resizing).
For PENet+, we report the representative activation setting (PReLU) identified in the $512{\times}512$ results.}}
\label{tab:acc_flops_256}
\small
\setlength{\tabcolsep}{6pt}
\renewcommand{\arraystretch}{1.2}
\begin{tabularx}{\textwidth}{l c c c *{6}{>{\centering\arraybackslash}X}}
\toprule
 & & & &
\multicolumn{2}{c}{\textbf{nsF5}} &
\multicolumn{2}{c}{\textbf{J-UNIWARD}} &
\multicolumn{2}{c}{\textbf{UERD}} \\
\cmidrule(lr){5-6}\cmidrule(lr){7-8}\cmidrule(lr){9-10}
\textbf{Model} &
$\mathbf{D_{\text{in}}}$ &
\textbf{Params (M)} &
\textbf{FLOPs (G)} &
\textbf{0.2 bpc} & \textbf{0.4 bpc} &
\textbf{0.2 bpc} & \textbf{0.4 bpc} &
\textbf{0.2 bpc} & \textbf{0.4 bpc} \\
\midrule
\rev{\textbf{PENet (baseline, ReLU)}} & \rev{5376} & \rev{16.16} & \rev{16.7}
& \rev{0.6653} & \rev{0.8601}
& \rev{0.5590} & \rev{0.7089}
& \rev{0.5572} & \rev{0.7002}
\\
\midrule
\multirow{3}{*}{\rev{\textbf{PENet+ (PReLU)}}}
& \rev{672}  & \rev{8.80} & \rev{0.33}
& \rev{0.6461} & \rev{0.8450}
& \rev{0.6510} & \rev{0.7874}
& \rev{0.7021} & \rev{0.8024}
\\
& \rev{1344} & \rev{9.98} & \rev{0.53}
& \rev{0.6583} & \rev{0.8396}
& \rev{0.7050} & \rev{0.8026}
& \rev{0.7087} & \rev{0.8425}
\\
& \rev{2688} & \rev{11.86} & \rev{1.26}
& \rev{0.6859} & \rev{0.8628}
& \rev{0.7631} & \rev{0.8629}
& \rev{0.7335} & \rev{0.8612}
\\
\bottomrule
\end{tabularx}
\end{table*}

\begin{revblock}
\subsubsection{Comparison with efficient baselines}
We additionally compare PENet+ with representative efficient steganalysis baselines (SRNet and ZhuNet) under the same dataset protocol and preprocessing at $512{\times}512$ and 0.4 bpc.
We report detection accuracy together with Params and FLOPs, using the same embedding settings and evaluation split as in Section~\ref{sec:exp-setup}.
Table~\ref{tab:light_baselines} summarizes the results.
\end{revblock}

\begin{table*}[t]
\centering
\caption{\rev{Comparison with efficient steganalysis baselines (ZhuNet and SRNet) on ALASKA2 QF90 at $512{\times}512$ and 0.4 bpc.
All methods are retrained and evaluated under the same dataset split and preprocessing.}}
\label{tab:light_baselines}
\small
\setlength{\tabcolsep}{6pt}
\renewcommand{\arraystretch}{1.15}
\begin{tabularx}{\textwidth}{l c c c c c >{\raggedright\arraybackslash}X}
\toprule
\textbf{Model} & \textbf{Params (M)} & \textbf{FLOPs (G)} &
\textbf{nsF5 (0.4 bpc)} & \textbf{J-UNIWARD (0.4 bpc)} & \textbf{UERD (0.4 bpc)} & \textbf{Note} \\
\midrule
\rev{ZhuNet~\cite{zhang2020depthwise}} &
\rev{2.87} & \rev{5.03} &
\rev{0.7266} & \rev{0.6035} & \rev{0.7158} &
\rev{Same protocol.}
\\

\rev{SRNet~\cite{srnet2018boroumand}} &
\rev{4.78} & \rev{24.29} &
\rev{0.8803} & \rev{0.8135} & \rev{0.7754} &
\rev{Same protocol.}
\\

\rev{PENet+ (ours)} &
\rev{8.80} & \rev{2.12} &
\rev{0.9070} & \rev{0.9040} & \rev{0.8613} &
\rev{Using PReLU.}
\\
\bottomrule
\end{tabularx}
\end{table*}

Although PENet+ has more parameters than ZhuNet and SRNet, its FLOPs are lower while its detection accuracy is consistently higher under the same protocol. Therefore, Table~\ref{tab:light_baselines} should be interpreted as an accuracy--computation trade-off rather than a pure parameter-efficiency comparison. PENet+ is not designed to be the smallest model in parameter count; instead, it targets high detection accuracy with substantially reduced forward computation.

\newcommand{\ModelColFont}{\small}  

\begin{figure*}[t]
\centering
\setlength{\tabcolsep}{2pt}
\renewcommand{\arraystretch}{1.0}

\begin{tabular}{@{}%
    >{\centering\arraybackslash\ModelColFont}m{0.18\textwidth}
    >{\centering\arraybackslash}m{0.25\textwidth}%
    >{\centering\arraybackslash}m{0.25\textwidth}%
    >{\centering\arraybackslash}m{0.25\textwidth}@{}}
\toprule
Model & (a) Stego & (b) Residual (Y) & (c) HPF activation energy \\
\midrule
PENet (v1, $K{=}62$) &
\includegraphics[width=\linewidth]{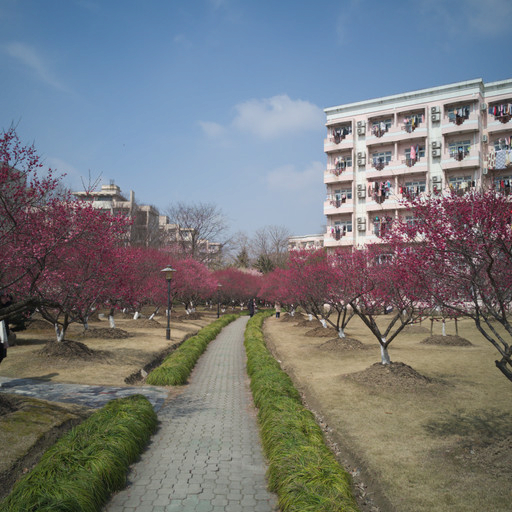} &
\includegraphics[width=\linewidth]{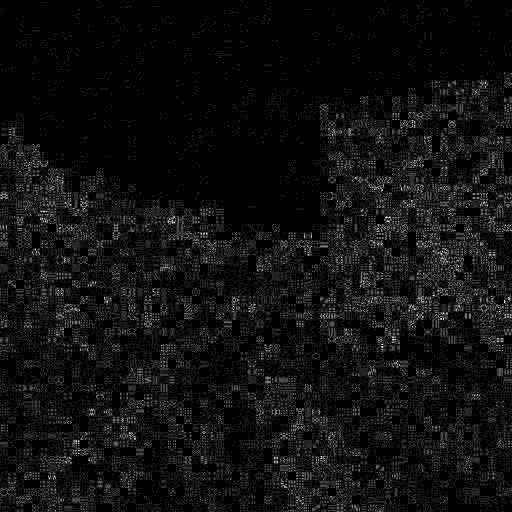} &
\includegraphics[width=\linewidth]{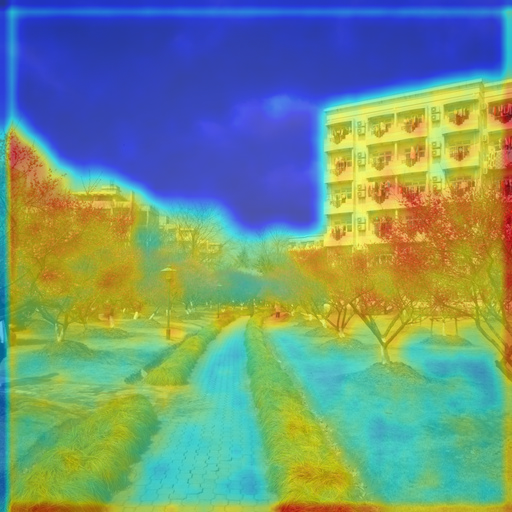} \\[3pt]

PENet+ (v3, $K{=}31$) &
\includegraphics[width=\linewidth]{stego/00142_stego.png} &
\includegraphics[width=\linewidth]{residual/00142_residual_y.png} &
\includegraphics[width=\linewidth]{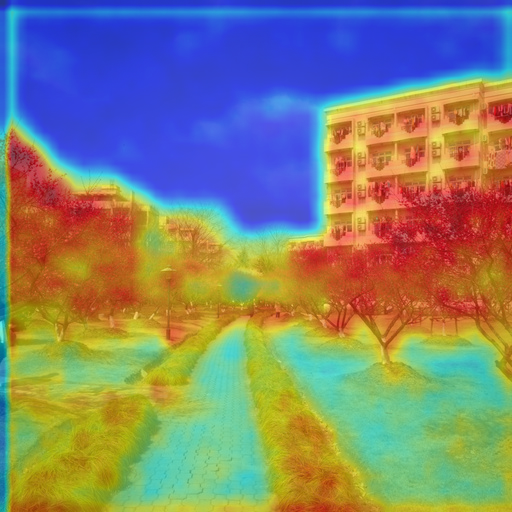} \\[3pt]

PENet+ (v2, $K{=}31$) &
\includegraphics[width=\linewidth]{stego/00142_stego.png} &
\includegraphics[width=\linewidth]{residual/00142_residual_y.png} &
\includegraphics[width=\linewidth]{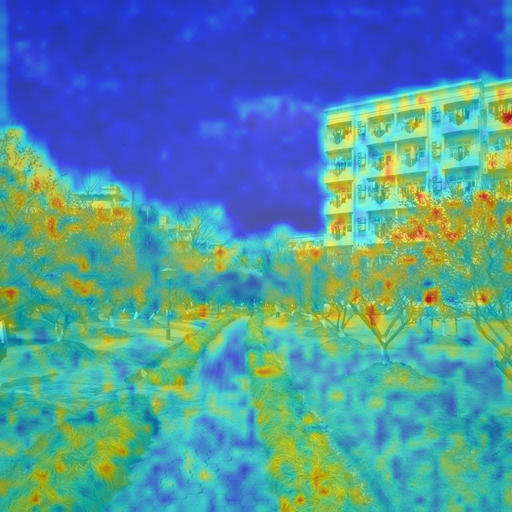} \\
\bottomrule
\end{tabular}

\caption{Class-agnostic HPF activation-energy overlays at the deepest backbone group for nsF5@0.4. Rows: original PENet using all 62 SRM--Gabor HPFs, PENet+ v3 with $K{=}31$ activation-based selected HPFs, and PENet+ v2 with $K{=}31$ activation-based selected HPFs. Columns: (a) stego image, (b) absolute residual $|\mathrm{cover}-\mathrm{stego}|$ on the Y channel, and (c) per-pixel activation-energy map aggregated over the selected HPF channels.}

\label{fig:hpf_activation_grid}
\end{figure*}

\begin{figure*}[t]
\centering
\setlength{\tabcolsep}{2pt}
\renewcommand{\arraystretch}{1.0}

\begin{tabular}{@{}%
    >{\centering\arraybackslash\ModelColFont}m{0.18\textwidth}
    >{\centering\arraybackslash}m{0.25\textwidth}
    >{\centering\arraybackslash}m{0.25\textwidth}@{}}
\toprule
Model & \shortstack[c]{(a) HPF activation energy\\(with ROI box)} & (b) Zoomed ROI\\
\midrule
PENet (v1, $K{=}62$) &
\includegraphics[width=\linewidth]{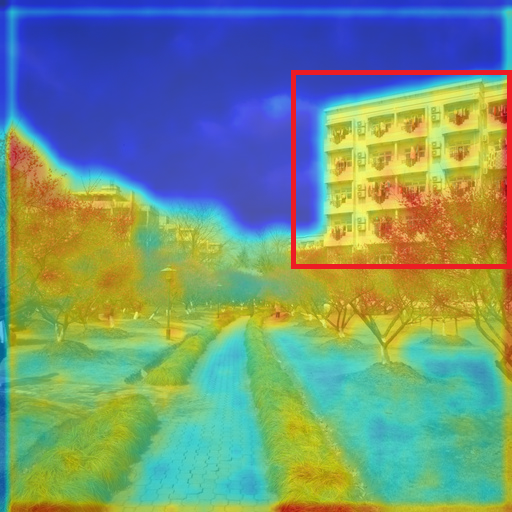} &
\includegraphics[width=\linewidth]{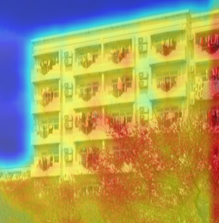} \\[3pt]

PENet+ (v3, $K{=}31$) &
\includegraphics[width=\linewidth]{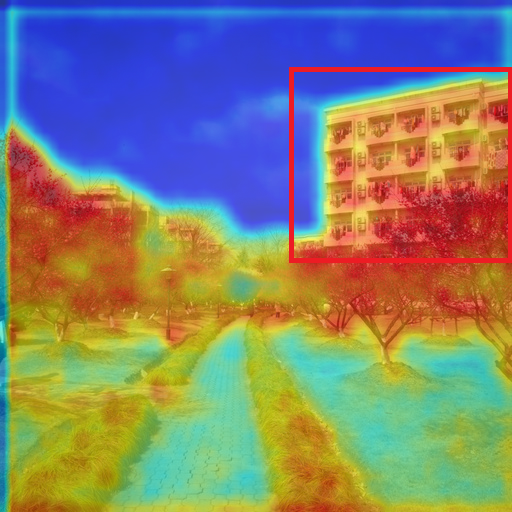} &
\includegraphics[width=\linewidth]{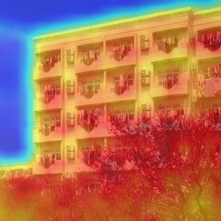} \\[3pt]

PENet+ (v2, $K{=}31$) &
\includegraphics[width=\linewidth]{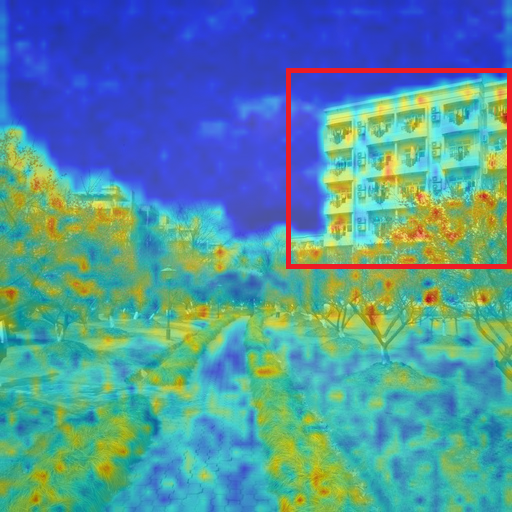} &
\includegraphics[width=\linewidth]{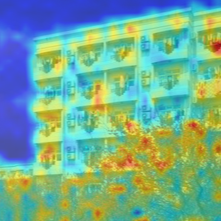} \\
\bottomrule
\end{tabular}

\caption{Zoomed comparison of class-agnostic HPF activation-energy maps at the deepest backbone group around the building region for nsF5@0.4. 
Rows: original PENet with 62 fixed HPFs, PENet+ v3 with $K{=}31$ activation-based selected HPFs, and PENet+ v2 with $K{=}31$ activation-based selected HPFs. 
Columns: (a) HPF activation map with the region of interest marked by a red box, and (b) zoomed view of the boxed region.}
\label{fig:hpf_activation_zoom}
\end{figure*}

Figure~\ref{fig:hpf_activation_grid} shows that the PReLU with HPFv3 variant produces activation maps whose high-response regions are tightly concentrated around the true modification patterns revealed by the residual map, whereas the original baseline tends to respond to broader background textures. This suggests that the proposed HPF selection and activation design help PENet+ focus its capacity on stego-prone regions rather than benign image content.

\begin{table*}[t]
\centering
\caption{Diagnostic comparison between v2 activation-aware pre-selection and v3 learnable selector across $D_{\text{in}}\in\{672,1344,2688\}$ at $512{\times}512$.
This table analyzes selector behavior under the same PReLU and SPP $\{4,2,1\}$ setting with HPF=31 (15 SRM + 16 Gabor), while the v3 rows are aligned with the final held-out PReLU results reported in Table~\ref{tab:acc_flops_full}.}
\label{tab:v2v3}
\small
\setlength{\tabcolsep}{8pt}
\renewcommand{\arraystretch}{1.2}
\begin{tabular}{l c c r c c c c}
\toprule
\textbf{Mode} & \textbf{HPF} & \textbf{SRM:Gabor} & $\mathbf{D_{\text{in}}}$ & \textbf{FLOPs (G)} &
\textbf{nsF5(@0.4)} & \textbf{J-UNIWARD(@0.4)} & \textbf{UERD(@0.4)} \\
\midrule
\multirow{3}{*}{v2 (pre-selection)} & \multirow{3}{*}{31} & \multirow{3}{*}{15:16}
&  672 & 2.12 & 0.9137  & 0.9023  & 0.8667  \\
&  &  & 1344 & 2.93 & 0.9214  & 0.9038  & 0.8642  \\
&  &  & 2688 & 5.85 & 0.9364  & 0.8463  & 0.8711  \\
\cmidrule(lr){1-8}
\multirow{3}{*}{v3 (learnable selector)} & \multirow{3}{*}{31} & \multirow{3}{*}{15:16}
&  672 & 2.12 & 0.9070  & 0.9040  & 0.8613  \\
&  &  & 1344 & 2.93 &0.9235  & \textbf{0.9356}  & \textbf{0.8850}  \\
&  &  & 2688 & 5.85 & \textbf{0.9481}  & 0.8552  & 0.8849  \\
\bottomrule
\end{tabular}
\vspace{-2mm}
\end{table*}

\begin{table}[t]
\centering
\caption{Module-wise parameters breakdown and total FLOPs on ALASKA2 dataset with QF90 ($512\times512$).}
\label{tab:complexity_breakdown}
\small
\setlength{\tabcolsep}{4.5pt}
\renewcommand{\arraystretch}{1.15}
\resizebox{\columnwidth}{!}{
\begin{tabular}{lcccccc}
\toprule
\multirow{2}{*}{Model} &
\multicolumn{5}{c}{Params (M)} &
\multirow{2}{*}{FLOPs (G)} \\
\cmidrule(lr){2-6}
& HPF & Backbone & Classifier & Others (MHSA) & Total & \\
\midrule
PENet (baseline)
& 0.00 & 2.13 & 5.64 & 8.39 & 16.16 & 67.48 \\

PENet+--L ($C{=}128$)
& 0.00 & 0.65 & 2.82 & 8.39 & 11.86 & 5.85 \\

PENet+--M ($C{=}64$)
& 0.00 & 0.17 & 1.41 & 8.38 & 9.98 & 2.93 \\

PENet+--S ($C{=}32$)
& 0.00 & 0.05 & 0.36 & 8.38 & 8.80 & 2.12 \\
\bottomrule
\end{tabular}
}
\end{table}

\begin{table}[t]
\centering
\caption{Ablation study results on the ALASKA2 dataset (QF=90, $512{\times}512$) with nsF5@0.4.}
\label{tab:ablation}
\begin{tabularx}{\linewidth}{l>{\centering\arraybackslash}X>{\centering\arraybackslash}X>{\centering\arraybackslash}X}
\toprule
\textbf{Variant} & \textbf{Accuracy} & \textbf{Params (M)} & \textbf{FLOPs (G)} \\
\midrule
A1: HPF 31 (pre-selection) &0.9257  &16.11  &51.27  \\
A2: + FC1 streamlining &0.9328  &11.74  &18.09  \\
A3: + Inverted Residual backbone &0.9278  &11.85  &5.85  \\
A4: + Activation (PReLU) &0.9481  &11.86  &5.85  \\
\bottomrule
\end{tabularx}
\end{table}
Table~\ref{tab:k_sweep} summarizes the effect of the HPF budget $K\!\in\!\{15,31,48\}$ under identical settings. On nsF5@0.4, $K{=}31$ attains the highest accuracy while avoiding the additional cost of $K{=}48$ (for example, PReLU reaches $0.9481$ at $K{=}31$ compared to $0.9402$ at $K{=}48$, whereas $K{=}15$ drops to $0.9214$). J-UNIWARD@0.4, $K{=}31$ is also the best configuration in our runs (PReLU $0.9356$), and UERD@0.4 follows the same pattern. These observations indicate that $K{=}31$ is a good operating point: it halves the original 62-channel HPF stem, preserves a balanced SRM–Gabor split, and, unlike $K{=}48$, avoids redundant filters that increase computation without improving accuracy.
To verify that the gain at $K{=}31$ does not simply come from using fewer filters, we further compare the proposed activation-based selection against a random baseline under the same HPF budget. For nsF5@0.4 with PReLU, we sample five random SRM--Gabor subsets for each $K\in\{15,31\}$ while preserving the same SRM--Gabor ratio, train PENet+ on each subset, and report the average accuracy in Table~\ref{tab:k_random}. The activation-based selection consistently outperforms random subsets, especially at the tighter budget $K{=}15$, which confirms that the activation-energy ranking provides informative guidance beyond random pruning.

Unless otherwise noted, subsequent experiments use activation-aware HPF selection with a balanced budget $K{=}31$ (15 SRM and 16 Gabor). We compare two modes: activation-aware pre-selection (v2) and a learnable selector (v3). For a fair comparison across architectures, we also sweep the classifier input width $D_{\text{in}}\!\in\!\{672,1344,2688\}$ at the SPP–FC1 interface. The corresponding results are reported in Table~\ref{tab:v2v3}.

\begin{revblock}
\noindent\textbf{Discussion and practical guidance.}
Table~\ref{tab:v2v3} suggests that the preferred selector can vary with the classifier input width $D_{\text{in}}$ and the embedding method.
With small $D_{\text{in}}$, the model has limited downstream capacity, making performance more sensitive to which residual channels are retained; in this regime, v2 can be more stable because it is based on direct activation-energy statistics.
As $D_{\text{in}}$ increases, the model can leverage richer residual combinations, and v3 may become beneficial by emphasizing filters aligned with embedding-specific artifacts (most notably for J-UNIWARD and UERD at the mid-width setting).
Accordingly, we recommend v2 for the smallest $D_{\text{in}}$ under tight compute budgets, v3 for the mid-width setting when targeting J-UNIWARD/UERD, and v2 as a reasonable default for the largest setting when simplicity and reproducibility are prioritized.
\end{revblock}

Because Table~\ref{tab:v2v3} is intended to compare selector behavior between v2 and v3, it should be read as a diagnostic selector analysis rather than as the main performance table. The v3 rows are aligned with the final held-out PReLU results in Table~\ref{tab:acc_flops_full}, while the main final results with statistical reporting are summarized in Table~\ref{tab:prelu_meanstd_512}.

\subsubsection{Complexity Analysis}

After HPF selection, an inverted-residual backbone, and classifier bottleneck optimization, PENet+ substantially reduces the computational cost of the original PENet while leaving the self-attention topology unchanged. Table~\ref{tab:complexity_breakdown} reports a module-wise parameter breakdown (HPF, convolutional backbone, classifier, and MHSA/Others) together with the total FLOPs per $512{\times}512$ image.

The original PENet contains 16.16\,M parameters and requires 67.48\,G FLOPs. Roughly half of the parameters reside in the MHSA/Others block, about one third in the classifier, and the rest in the convolutional backbone; the HPF stem is negligible in size. 
In contrast, the proposed PENet+ variants narrow both the backbone and the SPP$\rightarrow$FC1 interface while keeping the MHSA block structurally identical. The lightest configuration, \textbf{PENet+--S} with $C{=}32$ channels before SPP, uses 8.80\,M parameters and 2.12\,G FLOPs, corresponding to a \textbf{45.5\%} reduction in parameters and a \textbf{96.9\%} reduction in FLOPs compared with PENet. The intermediate variants, PENet+--M ($C{=}64$) and PENet+--L ($C{=}128$), offer accuracy--efficiency trade-offs with 9.98\,M / 2.93\,G and 11.86\,M / 5.85\,G, respectively.

An internal profiling of the forward pass shows that high-resolution convolution and attention dominate the overall cost: the convolutional backbone and MHSA/Others together account for more than 95\% of total FLOPs, whereas the HPF stem and classifier each contribute less than 1\%. This confirms that the majority of the computational burden in PENet arises from high-resolution feature processing, and that PENet+ effectively reduces this computational burden through an inverted-residual backbone and a narrow classifier bottleneck without changing the original attention topology.

\subsection{Ablation Study}

To analyze the contribution of each component in PENet+, we progressively add or replace modules from the original PENet baseline.
All ablation experiments are performed on the ALASKA2 QF90 dataset at $512{\times}512$ resolution to evaluate performance under high computational demand.

\begin{itemize}
    \item \textbf{(1) HPF Pre-Selection.} 
    The number of high-pass filters is reduced from 62 to 31 using activation-energy ranking and balanced SRM--Gabor selection,
    preserving filter diversity while halving early feature-map computation.

    \item \textbf{(2) Inverted Residual Backbone.} 
    The original PDConv stacks are replaced with MobileNetV2-style inverted residual (IR) blocks to improve spatial efficiency and maintain residual sensitivity through lightweight expansion--depthwise--projection operations.

    \item \textbf{(3) Activation Function.} 
    We evaluate five activation functions 
    (\textbf{ReLU}, \textbf{ReLU6}, \textbf{LeakyReLU}, \textbf{SiLU}, and \textbf{PReLU}) under identical conditions to compare their effects on convergence and detection accuracy.

    \item \textbf{(4) FC1 Streamlining.} 
    The classifier input dimension after SPP is reduced (e.g., $5376\!\rightarrow\!672$) by narrowing channels before SPP, significantly lowering parameter count and FLOPs while maintaining comparable classification performance.
\end{itemize}

We denote incremental configurations as A1--A4, where A1 applies (1) only, and A4 represents the full PENet+ configuration with all components.
Each variant is trained and evaluated under identical optimization settings, allowing clear observation of individual and cumulative contributions.

\label{sec:ablation-results}

We perform a stepwise ablation from A1 to A4 as defined in Section~\ref{sec:ablation-results}.
Starting from the original PENet baseline, A1 replaces the 62 SRM+Gabor filters with 31 pre-selected HPFs. A2 streamlines the first fully connected layer to reduce parameters and FLOPs. A3 introduces an inverted residual backbone as the full lightweight architecture. Finally, A4 adopts PReLU activations, which further improves detection accuracy.
Table~\ref{tab:ablation} summarizes the progressive effects of each modification.
It should be noted that this ablation is a progressive design study rather than a strict single-factor ablation. In particular, the transition from A2 to A3 changes both the backbone operator and the effective channel width, because the inverted residual backbone is introduced together with the lightweight channel configuration. Therefore, A3 is intended to show the cumulative effect of moving to the lightweight backbone design rather than the isolated effect of a single operator replacement. The transition from A3 to A4 keeps the FLOPs nearly unchanged but improves accuracy, indicating that PReLU is especially beneficial under compressed residual representations. We attribute this improvement to its ability to preserve weak negative residual responses that ReLU-type activations may suppress, which is important for detecting subtle stego artifacts.

\subsection{Visualization and Discussion}
\label{sec:visualization}

We analyze how PENet+ localizes stego-prone regions using the visualization results in Figures~\ref{fig:hpf_activation_grid} and \ref{fig:hpf_activation_zoom}.
Figure~\ref{fig:hpf_activation_grid} compares the original PENet with the proposed PENet+ (v2 and v3) by showing the stego image, its Y–channel absolute residual map, and the class-agnostic HPF activation-energy map at the deepest backbone group.
The PENet+ variants, especially the learnable selection (v3), produce activation responses that are more spatially concentrated around the true modification regions revealed in the residual map, whereas the original PENet often highlights broader background textures.

To further inspect local behavior, Figure~\ref{fig:hpf_activation_zoom} provides a zoomed comparison around the building area.
The ROI box (left column) and its magnified region (right column) show that PENet+ suppresses irrelevant background activation and sharpens responses on the actual stego-affected structures.
This indicates that activation-aware HPF selection and lightweight backbone design do not merely reduce computation but also guide the network to retain stego-discriminative cues more effectively.

\section{Conclusion}
In this paper, we introduced PENet+, a lightweight steganalysis framework that preserves the discriminative strength of PENet while greatly reducing computational cost. Rather than redesigning the self-attention structure, we improved efficiency through three complementary components: a learnable high-pass filter selection module that derives a compact SRM--Gabor subset from activation statistics, a MobileNetV2-style inverted residual backbone for efficient spatial modeling, and a streamlined classifier that reduces the SPP$\rightarrow$FC1 bottleneck. 
Our results show that the proposed HPF design converges stably and retains an informative filter bank despite using far fewer filters than the original PENet. We also found that PReLU is more effective than ReLU-based alternatives because it preserves weak negative residual responses that are useful for detecting subtle stego traces. 
Experiments on our disjoint ALASKA2 JPEG QF90 protocol at $512{\times}512$ show that PENet+ reduces parameters by about 45.5\% and FLOPs by about 97\% compared with the re-evaluated PENet baseline, while maintaining competitive detection accuracy across multiple steganographic methods. These findings suggest that efficient Transformer-based steganalysis can substantially reduce computational overhead, although actual device-level latency and power measurements remain an important direction for future work.


\section*{Acknowledgment}
This work was supported by the Institute of Information $\&$ Communications Technology Planning $\&$ Evaluation (IITP) grant funded by the Korea government (MSIT) [RS-2021-II211341, Artificial Intelligence Graduate School Program (Chung-Ang University) and RS-2022-II220124, Development of Artificial Intelligence Technology for Self-Improving Competency-Aware Learning Capabilities]. SNUAILAB, corp, also supports this work.

{\small
\bibliographystyle{IEEEtran}
\bibliography{egbib}
}
\begin{IEEEbiography}[{\includegraphics[width=1in,height=1.25in,clip,keepaspectratio]{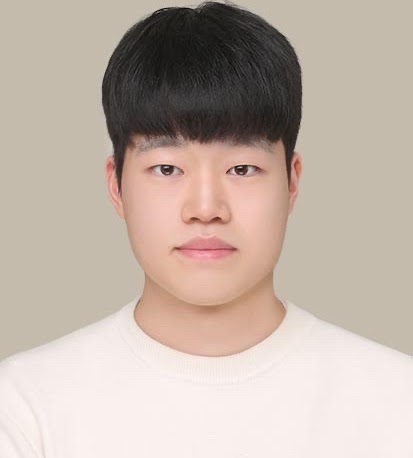}}]{Jincheol An} received the B.S. degree in mechanical engineering from Chung-Ang University, Seoul, South Korea, in 2025, where he is currently pursuing the M.S. degree in Artificial Intelligence. His research interests include image steganalysis, watermarking and deep learning-based security systems.
\end{IEEEbiography}

\begin{IEEEbiography}[{\includegraphics[width=1in,height=1.25in,clip,keepaspectratio]{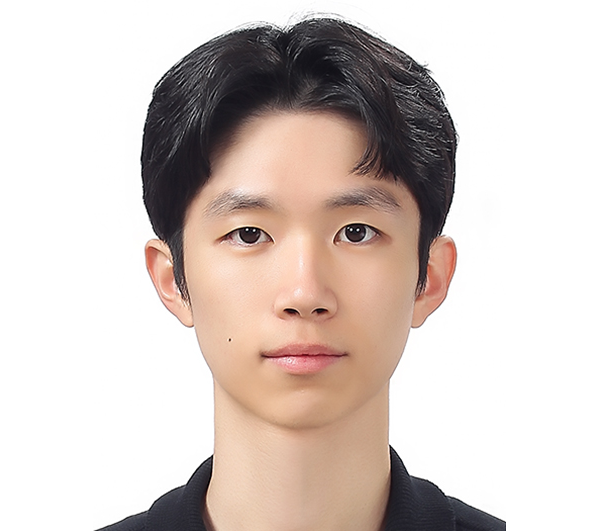}}]{Dongsu Kim} received the B.S. degree in computer engineering from Hanbat National University, Daejeon, South Korea, in 2025, where he is currently pursuing the M.S. degree in computer engineering. His research interests include image steganalysis, deep learning-based security systems, and computer vision applications.
\end{IEEEbiography}

\begin{IEEEbiography}[{\includegraphics[width=1in,height=1.25in,clip,keepaspectratio]{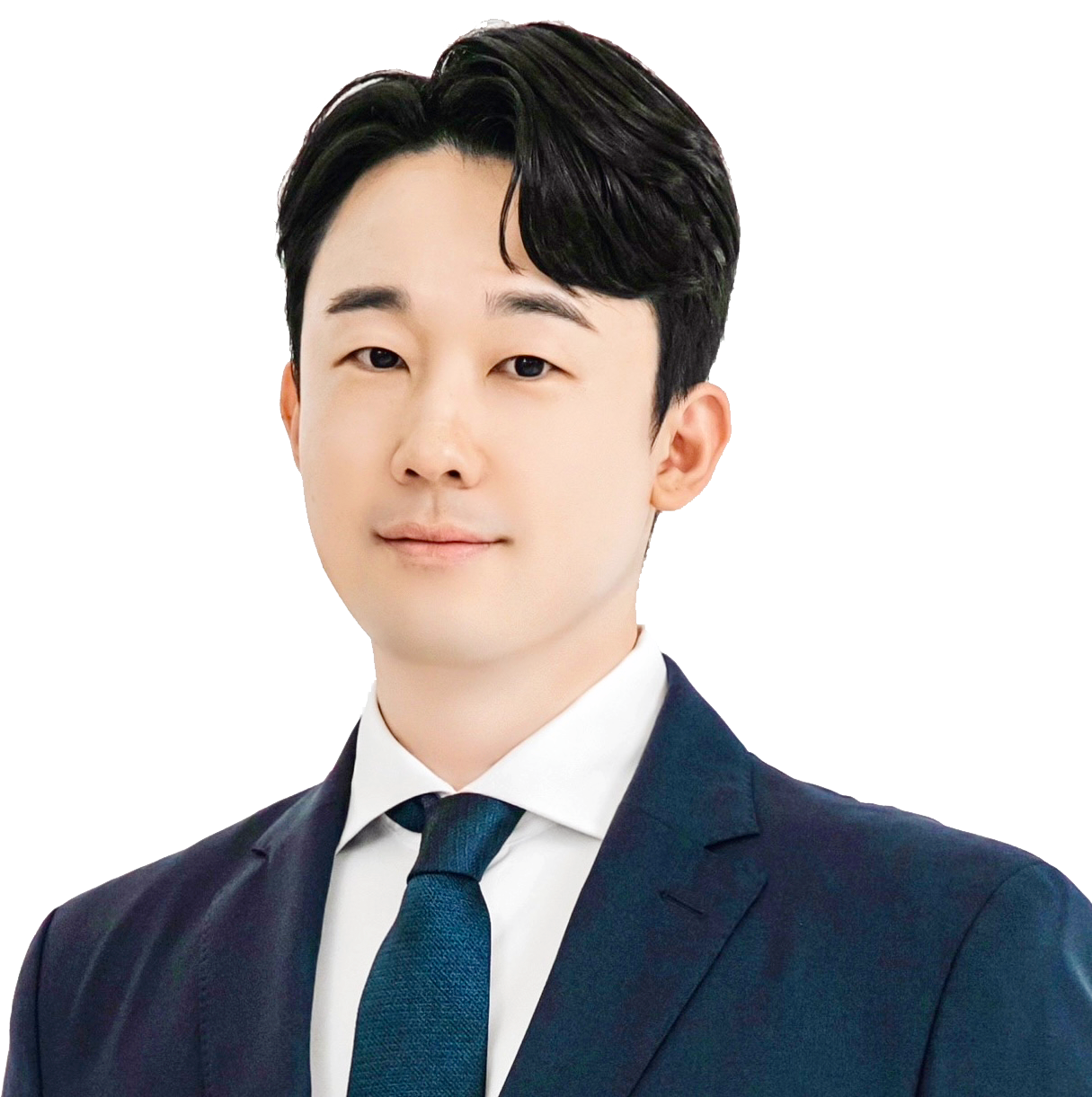}}]{Haneol Jang} received the B.E. degree in information and computer engineering from Ajou University, Korea, the M.S. degree in computer science from KAIST, Korea, and the Ph.D. degree in computer science from KAIST, Korea, in 2012, 2014, and 2018, respectively. He is currently an Associate Professor in the Department of Computer Engineering at Hanbat National University. Prior to his academic appointment, he was an AI researcher at NAVER in 2018, where he developed image watermarking technology for webtoon content. From 2018 to 2020, he served as a Senior Researcher at National Security Research, specializing in AI security technologies. His research interests include multimedia security, remote sensing image processing, computer vision, and multimodal AI.
\end{IEEEbiography}

\begin{IEEEbiography}[{\includegraphics[width=1in,height=1.25in,clip,keepaspectratio]{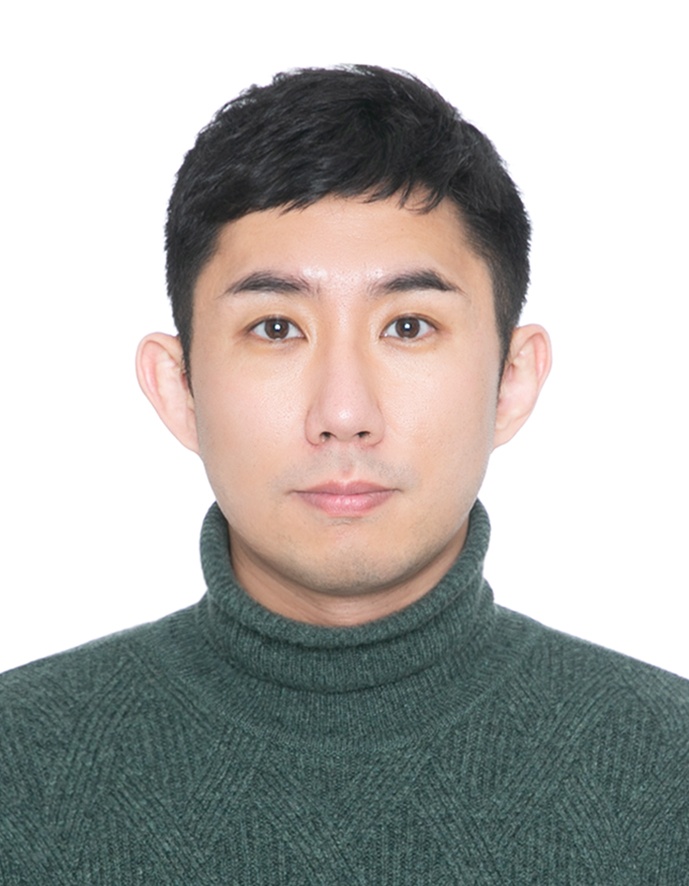}}]{Youngjoon Yoo}
received the B.S. degree in Electrical and Computer Engineering from Seoul National
University, Seoul, Korea, in 2011, and the Ph.D. degree in Electrical Engineering from the same university in 2017.
He was a research scientist in NAVER AI Research and led the Image Vision team at NAVER CLOVA. 
He is currently an Assistant Professor with the Department of Artificial Intelligence, Chung-Ang University, Seoul, Korea.
His research interests include deep learning for computer vision and probabilistic modeling.
\end{IEEEbiography}

\EOD
\end{document}